# SustainDC: Benchmarking for Sustainable Data Center Control

**Avisek Naug[†], Antonio Guillen[†], Ricardo Luna[†], Vineet Gundecha[†],**
**Desik Rengarajan, Sahand Ghorbanpour, Sajad Mousavi, Ashwin Ramesh Babu,**
**Dejan Markovikj, Lekhapriya D Kashyap, Soumyendu Sarkar[†*]**

Hewlett Packard Enterprise (Hewlett Packard Labs)

{avisek.naug, antonio.guillen, rluna, vineet.gundecha, desik.rengarajan,
sahand.ghorbanpour, sajad.mousavi, ashwin.ramesh-babu, dejan.markovikj,
lekhapriya.dheeraj-kashyap, soumyendu.sarkar}@hpe.com

## Abstract

Machine learning has driven an exponential increase in computational demand, leading to massive data centers that consume significant amounts of energy and contribute to climate change. This makes sustainable data center control a priority. In this paper, we introduce SustainDC, a set of Python environments for benchmarking multi-agent reinforcement learning (MARL) algorithms for data centers (DC). SustainDC supports custom DC configurations and tasks such as workload scheduling, cooling optimization, and auxiliary battery management, with multiple agents managing these operations while accounting for the effects of each other. We evaluate various MARL algorithms on SustainDC, showing their performance across diverse DC designs, locations, weather conditions, grid carbon intensity, and workload requirements. Our results highlight significant opportunities for improvement of data center operations using MARL algorithms. Given the increasing use of DC due to AI, SustainDC provides a crucial platform for the development and benchmarking of advanced algorithms essential for achieving sustainable computing and addressing other heterogeneous real-world challenges.

## 1 Introduction

One of the growing areas of energy and carbon footprint ($CFP$) can be traced to cloud data centers (DCs). The increased use of cloud resources for batch workloads related to AI model training, multimodal data storage and processing, or interactive workloads like streaming services, hosting websites have prompted enterprise clients to construct numerous data centers. Governments and regulatory bodies are increasingly focusing on environmental sustainability and imposing stricter regulations to reduce carbon emissions. This has prompted industry-wide initiatives to adopt more intelligent DC control approaches. This paper presents SustainDC, a sustainable DC Multi-Agent Reinforcement Learning (MARL) set of environments. SustainDC helps promote and prioritize sustainability and facilitates an ecosystem of AI researchers with a platform to contribute to a more environmentally responsible DC.

The main contributions of this paper are the following:

- A highly customizable collection of environments related to Data Center (DC) operation that can be used to benchmark energy and carbon footprint for different DC designs. The ability to subclass models across the spectrum of DC components like workloads, specification of individual servers, to cooling towers, allows the user to test fine-grained design choices.

---

*Corresponding author. †These authors contributed equally.

- The environments are wrapped in the Gymnasium *Env* class, lending it itself to benchmarking of different control strategies for optimizing energy, carbon footprint, and related metrics.

- Supports MARL controllers with both homogeneous and heterogeneous agents, as well as non-ML controllers. With the given environments, we undertake extensive studies to show the benefits and disadvantages of a collection of multi-agent approaches.

- SustainDC allows the user to perform reward shaping to help run ablation studies on different parts of the DC for performance optimization of the different areas.

Code, licenses, and instructions of SustainDC can be found on GitHub[2]. Documentation can be found here [3].

## 2 Related Work

Recent advancements in the field of Reinforcement Learning (RL) have led to an increased focus on optimizing energy consumption in areas such as building and DC management. This has resulted in the development of several environments for RL applications. *CityLearn* [Vázquez-Canteli et al., 2019] is an open-source platform that supports single and MARL strategies for energy coordination and demand response in urban environments. *Energym* [Scharnhorst et al., 2021], *RL-Testbed* [Moriyama et al., 2018] and *Sinergym* [Jiménez-Raboso et al., 2021] were developed as RL wrappers that facilitate communication between Python and EnergyPlus, enabling RL evaluation on the collection of buildings modeled in EnergyPlus. *SustainGym* Yeh et al. [2023] is one of the latest suite of general purpose RL tasks for evaluation of sustainability, simulating electric vehicle charging scheduling and battery storage bidding in electricity markets.

Most of the above-mentioned works use *EnergyPlus* [Crawley et al., 2000] or, *Modelica* Wetter et al. [2014], Zuo et al. [2021] which were primarily designed for modeling thermo-fluid interactions with traditional analytic control with little focus on Deep Learning applications. The APIs provided in these works only allow sampling actions in a model free manner, lacking an easy approach to customization or re-parameterization of system behavior. This is due to the fact that most of the works have a set of pre-compiled binaries (e.g. FMUs in Modelica) or fine-tuned spline functions (in EnergyPlus) to simulate nominal behavior. Furthermore, there is a significant bottleneck in using these precompiled environments from Energyplus or Modelica for Python based RL applications due to latency associated with cross-platform interactions, versioning issues in traditional compilers for EnergyPlus and Modelica, unavailability of open source compilers and libraries for executing certain applications.

SustainDC allows users to simulate the electrical and thermo-fluid behavior of large DCs directly in Python. Unlike other environments that rely on precompiled binaries or external tools, SustainDC is easily end-user customizable and fast It enables the design, configuration, and control benchmarking of DCs with a focus on sustainability. This provides the ML community with a new benchmark environments specifically for Heterogeneous MARL in the context of DC operations, allowing for extensive goal-oriented customization of the MDP transition function, state space, actions space, and rewards.

## 3 Data Center Operational Model

Figure 1 shows the typical components of a DC operation modeled in SustainDC. *Workloads* are uploaded to the DC from a proxy client. For non-interactive batch workloads, a fraction of these jobs can be flexible or delayed to different time periods. This creates a scheduling problem where workloads can be delayed to some time of the day when *Grid Carbon Intensity* (*CI*) is lower.

Next, as the servers (IT system) in the DC process these workloads, they generate heat, which needs to be removed. Hence, a complex HVAC system comprising multiple components is set up to cool the IT system. As shown in Figure 2, the warm air leaves the servers and rises up by convection. Due

---

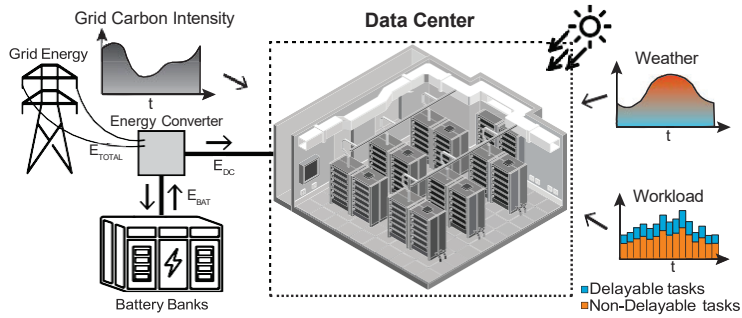

Figure 1: Operational Model of a SustainDC Data Center

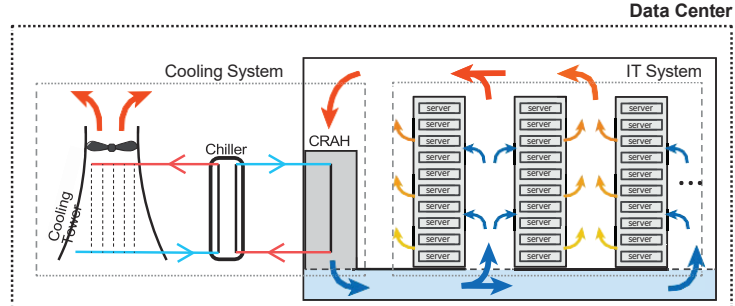

Figure 2: Model of the data center. The configuration allows customization of the number of cabinets per row, the number of rows, and the number of servers per cabinet. The cooling system, comprising the CRAH, chiller, and cooling tower, manages the heat generated by the IT system.

to the forced draft of the HVAC fan, this warm air enters the *Computer Room Air Handler* (CRAH) (shown using red arrows) where it is cooled down to an optimal setpoint by a heat exchange process using a "primary" chilled water loop. The chilled air is then sent back to the IT room using a plenum located underneath the DC (shown using blue arrows). The warm water then goes back to the *Chiller* where this heat is transferred by another heat exchange process to the "secondary" chilled water loop that carries the heat to a *Cooling Tower*. The cooling tower fan operates at a certain speed to reject the heat from the secondary chilled water to the outside environment. This speed, and hence the energy consumption of the fan, is a function of the inlet temperature of the secondary chilled water loop at the cooling tower, the required setpoint for the outlet temperature, the outside air temperature, and humidity. Thus, depending on the external *Weather* and *Workload* processed, the IT and cooling system consume *Grid Energy*. Choosing the optimal cooling setpoint for the CRAH has the potential to reduce the carbon footprint of the DC. Furthermore, the optimal cooling setpoint also influences the energy efficiency of the servers (Sun et al. [2021]).

Often, larger DCs include onsite *Battery Banks*. Batteries can be charged from the grid during low *CI* periods. During higher *CI* periods, they provide auxiliary energy to the DC. This creates an opportunity to solve a decision problem where we optimally choose the time intervals during which the battery charges versus when it is used to provide auxiliary energy.

We observed that these three control problems are related in a cascading manner. This motivates the development of testbeds or environments where we can test different multiagent control approaches to reduce the carbon footprint and other associated sustainability metrics of interest.

## 4   SustainDC environment overview

We provide a high-level overview of the SustainDC in Figure 3. We identify the three main environments developed in *Python* and highlight their individual components, customization capabilities, and associated control problems. The *Workload Environment* models and controls the execution and scheduling of delay-tolerant workloads within the DC. Inside the *Data Center Environment*, the servers housed in the IT room cabinets process these workloads. It simulates the electric and thermal-fluid behavior, and the resulting heat generated from running these jobs is conducted to the outside environment by means of a set of HVAC cooling components. The *Battery Environment*

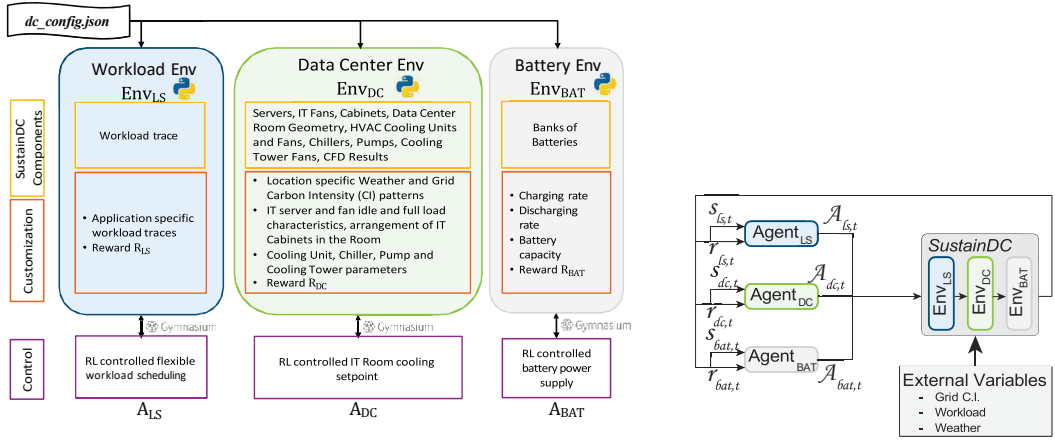

(a) High-level overview of SustainDC, showing the three main environments (*Workload Env*, *Data Center Env*, and *Battery Env*) along with their customizable components and control actions.

(b) RL loop in SustainDC, depicting how states and actions are formed from individual agents.

Figure 3: SustainDC overview and RL loop

simulates charging from the grid during off-peak hours and provides auxiliary energy to the DC during peak grid carbon intensity periods. The detailed physics-based implementation of the individual environments is provided in the supplemental document. The customization can be completely specified via *dc_config.json*, which is the universal configuration file for specifying every aspect of the DC environment design in SustainDC.

Figure 3a shows an overview of SustainDC, highlighting the *Workload Environment*, *Data Center Environment*, and *Battery Environment* with their customizable parameters. Figure 3b depicts the RL loop in SustainDC, illustrating how agents' actions and states optimize DC operations, considering external variables like grid CI, workload, and weather.

## 4.1 Workload Environment

The *Workload Environment* ($Env_{LS}$) manages the execution and scheduling of delayable workloads within the DC. It does this by streaming workload traces (measured in FLOPs of compute) over a specified period. SustainDC includes a collection of open-source workload traces from *Alibaba* Alibaba Group [2017] and *Google* Google [2019] data centers. Users can customize this component by adding new workload traces to the appropriate folder (*data/Workload*) or specifying a path to existing traces in the *dc_config.json*.

Certain workloads are flexible, meaning they can be rescheduled within an allowable time horizon. These tasks, such as update tasks or backup tasks, do not need to be executed immediately and can be delayed based on their urgency or Service-Level Agreement (SLA). This flexibility allows the workload to be shifted to periods when the grid's CI is lower, thereby reducing the DC's overall carbon footprint (*CFP*).

Users can also customize the CI data. By default, we provide a one-year CI file for the following states: Arizona, California, Georgia, Illinois, New York, Texas, Virginia, and Washington. These locations were selected because they are where most data centers are situated. The carbon intensity files are extracted from eia.gov (https://api.eia.gov/bulk/EBA.zip) and located in the folder *data/CarbonIntensity*.

Let $B_t$ be the instantaneous DC workload trace at time $t$, with $X\%$ of the load being rescheduled up to $N$ simulation steps into the future. The goal of an RL agent ($Agent_{LS}$) is to observe the current time of day ($SC_t$), the current and forecast grid CI data ($CI_{t...t+L}$), and the amount of rescheduled workload left ($D_t$). Based on these observations, the agent decides an action $A_{ls,t}$ (as shown in Table 1) to reschedule the flexible component of $B_t$, minimizing the net *CFP* over $N$ steps.

## 4.2 Data Center Environment

The *Data Center* environment ($Env_{DC}$) includes an exhaustive set of models and associated specifications that can be configured. For IT level design, SustainDC allows users to define the dimensions of the IT Room, the arrangement of the server cabinets (including the number of *rows* and *cabinets* per row), and the *approach* and *return* temperatures. Users can also specify the power characteristics of the servers and fans, such as *idle power*, *rated full load power*, and *rated full load frequency*.

On the cooling side, SustainDC allows customization of the *chiller reference power*, *cooling fan reference power*, and the *setpoint* for the supply air used to cool the IT Room. It also includes specifications for the pump and cooling tower characteristics, such as *rated full load power* and *rated full load frequency*. All these parameters can be configured in the *dc_config.json* file.

One of the key advantages of SustainDC is its ability to automatically adjust the cooling capacities for the HVAC components based on workload requirements and IT room specifications. This process, known as "sizing," ensures that the DC is neither over-cooled nor over-heated during operation. Previous environments did not offer this capability, often leading to inaccurate results. For instance, changing the IT room configuration in other environments only affected IT energy consumption without considering the overall cooling requirements, leading to inconsistent RL-based control outcomes, e.g.: in *RL-Testbed* in Moriyama et al. [2018]. SustainDC addresses this by integrating custom supply and approach temperatures derived from Computational Fluid Dynamics (CFD) simulations, simplifying the complex calculations of temperature changes between the IT Room HVAC and the IT Cabinets Sun et al. [2021].

In addition, SustainDC includes weather data (in *data/Weather*) in the .epw format for the same locations as the CI data. This data, sourced from https://energyplus.net/weather, represents typical weather conditions for these regions. Users can also specify their own weather files if needed.

If $\hat{B}_t$ is the adjusted workload from the *Workload Environment*, the goal of the RL agent ($Agent_{DC}$) would be to choose an optimal cooling setpoint $A_{dc,t}$ (Table 1) such that the net carbon footprint *CFP* from the resulting cooling ($E_{hvac}$) and IT ($E_{it}$) energy consumptions over an N step horizon is minimum. In SustainDC, by default, the agent state space comprises the time of day and year ($SC_t$), the ambient weather ($t_{db}$), the IT Room temperature ($t_{room}$), the previous step cooling ($E_{hvac}$) and IT ($E_{it}$) energy consumptions, and the forecast grid CI data ($CI_{t...t+L}$).

## 4.3 Battery Environment

The *Battery Environment* ($Env_{BAT}$) is based on models for charging and discharging batteries, such as $f_{charging}(BatSoc, \delta\tau)$ from Acun et al. [2023b]. Parameters for these components, as well as the battery capacity, can be specified in *dc_config.json*.

The goal of the RL agent ($Agent_{BAT}$) is to manage the battery's state of charge ($BatSoc_t$) effectively. Based on the net energy consumption ($E_{hvac} + E_{it}$) from the *Data Center* environment, the time of day ($SC_t$), the battery's state of charge ($BatSoc_t$), and the forecast grid CI data ($CI_{t...t+L}$), the agent must decide on an action $A_{bat,t}$ (as shown in Table 1). The actions include whether to charge the battery from the grid, do nothing, or provide auxiliary energy to the data center, all aimed at minimizing the overall carbon footprint.

## 4.4 Heterogeneous Multi Agent Control Problem

While SustainDC allows the user to solve the individual control problems for the three environments, the goal of the paper is to establish a multi-agent control benchmark which allows us to jointly optimize the *CFP* by considering the joint actions of all the three agents ($Agent_{LS}$, $Agent_{DC}$, and $Agent_{BAT}$). The sequence of operations for the joint multi-agent and Multi-Environment functions can be represented as:

$$Agent_{LS} : (SC_t \times CI_t \times D_t \times B_t) \rightarrow A_{ls,t} \tag{1}$$

$$Agent_{DC} : (SC_t \times t_{db} \times t_{room} \times E_{hvac} \times E_{it} \times CI_t) \rightarrow A_{dc,t} \tag{2}$$

$$Agent_{BAT} : (SC_t \times Bat\_SoC \times CI_t) \rightarrow A_{bat,t} \tag{3}$$

$$Env_{LS} : (B_t \times A_{ls,t}) \rightarrow \hat{B}_t \tag{4}$$

$$Env_{DC} : (\hat{B}_t \times t_{db} \times t_{room} \times A_{dc,t}) \rightarrow (E_{hvac}, E_{it}) \tag{5}$$

$$Env_{BAT} : (Bat\_SoC \times A_{bat,t}) \rightarrow (Bat\_SoC, E_{bat}) \tag{6}$$

$$CFP_t = (E_{hvac} + E_{it} + E_{bat}) \times CI_t \tag{7}$$

where $E_{bat}$ is the net discharge from the battery based on the change in battery SoC $Bat\_SoC$. It can be positive or negative depending on the action $A_{bat,t}$. In case it is providing auxiliary energy, $E_{bat}$ is negative. If it is charging from the grid, $E_{bat}$ is a positive value.

The goal of the multi-agent problem is to find $\theta_{LS}$, $\theta_{DC}$ and $\theta_{BAT}$ that parameterize the corresponding policies $Agent_{LS}$, $Agent_{DC}$ and $Agent_{BAT}$ , such that net $CFP$ is minimized over a specified horizon N. Here we choose N to be $(31 \times 24 \times 4)$ i.e. a horizon of 31 days, where we assume a step duration of 15 minutes.

$$\left( \theta_{LS}, \theta_{DC}, \theta_{BAT} \right)^{-} = argmin \sum_{t=0}^{t=N} \overline{CFP_t} \tag{8}$$

### 4.5 Rewards

While $CFP$ reduction is the default objective in SustainDC, the reward formulation is highly customizable and allows users to consider other objectives, including total energy usage and total operating cost across all the DC components and water usage. Primarily, we consider the following default rewards for the three environments ($Env_{LS}$, $Env_{DC}$, $Env_{BAT}$ )

$$(r_{LS}, r_{DC}, r_{BAT}) = \left( -(CFP_t + LS_{Penalty}), -(E_{hvac,t} + E_{it,t}), -(CFP_t) \right)^{-}$$

Here, $LS_{Penalty}$ is a penalty attributed to the Load Shifting Agent ($Agent_{LS}$) in the Workload Environment ($Env_{LS}$) if it fails to assign the flexible workloads that were supposed to be rescheduled within the time horizon $N$ . This implies that if $D_t$ is positive at the end of a horizon $N$ , we assign $LS_{Penalty}$. The details for evaluating $LS_{Penalty}$ is discussed in the supplemental document. The user can choose to use any other reward formulation by subclassing the base reward class inside *utils/reward_creator.py*.

| Agent | Control Knob | Actions | | Optimization Strategy | Figure |
|---|---|---|---|---|---|
| **$Agent_{LS}$** | Delayable workload scheduling | 0 <br> 1 <br> 2 | Store Delayable Tasks <br> Compute All Immediate Tasks <br> Maximize Throughput | Shift tasks to periods of lower CI/lower external temperature/other variables to reduce the $CFP$ . | Workload <br><br> — Orig. Delayable work. <br> ⋯ Delayed work. shifted <br> ■ Carbon Intensity |
| **$Agent_{DC}$** | Cooling Setpoint | 0 <br> 1 <br> 2 | Decrease Setpoint <br> Maintain Setpoint <br> Increase Setpoint | Optimize cooling by adjusting cooling setpoints based on workload, external temperature, and CI. | Setpoint <br><br> —External Temperature <br> —Cooling Setpoint <br> ■ Carbon Intensity |
| **$Agent_{BAT}$** | Battery energy supply/store | 0 <br> 1 <br> 2 | Charge Battery <br> Hold Energy <br> Discharge Battery | Store energy when CI/temperature/workload/other is low and use stored energy when is high to reduce $CFP$. | Battery <br><br> — Charging  Discharging |

Table 1: Overview of control choices in SustainDC: the tunable knobs, the respective action choices, optimization strategies, and visual representations.

Based on the individual rewards, we can formulate an independent or a collaborative reward structure where each agent gets partial feedback in the form of rewards from the other agent-environment pair. The collaborative feedback reward formulation for each agent is formulated as:

$$R_{LS} = a * r_{LS} + (1 - a)/2 * r_{DC} + (1 - a)/2 * r_{BAT}$$
$$R_{DC} = (1 - a)/2 * r_{LS} + a * r_{DC} + (1 - a)/2 * r_{BAT}$$
$$R_{BAT} = (1 - a)/2 * r_{LS} + (1 - a)/2 * r_{DC} + a * r_{BAT}$$

Here $a$ is the weighting parameter. The reward-sharing mechanism allows the agents to estimate the feedback from their actions in other environments for independent critic multiagent RL algorithms (e.g. IPPO de Witt et al. [2020]). For example, the adjusted CPU Load $\hat{B}_t$ affects the data center energy demand, $E_{cool} + E_{it}$ which in turn affects the battery optimizer's decision to charge or discharge resulting in a particular net $CO_2$ *Footprint*. Hence, we decided to investigate a collaborative reward structure. We conduct a set of ablation experiments with different values of $a$ to understand whether performance can be via reward sharing mechanism.

# 5 Evaluation Metrics and Experimental Settings

We consider five metrics when evaluating different RL approaches on SustainDC. *$CO_2$ footprint* ( *CFP* ) indicates the cumulative carbon footprint of the DC operation over the period of evaluation. *HVAC Energy* refers to the amount of energy consumed across the DC cooling components including the chiller, pumps and cooling tower. *IT energy* refers to the amount of energy consumed across the DC servers. *Water Usage* refers to the chilled water that is recirculated through the cooling system. In certain DCs, the availability of chilled water from a central plant is limited, and efficient utilization of this resource helps lower the water footprint of the DC. *Task Queue*: In our approach, since we reschedule workloads over time, this metric keeps track of how many FLOPs of compute are accumulating that need to be rescheduled at a later time period under lower CI. Higher values of Task Queue indicate worse SLAs in DCs.

We ran the experiments on an Intel(R) Xeon(R) Platinum 8470 server with 104 CPUs using 4 threads per training agent. All the hyperparameter settings for the benchmark experiments are provided in the supplemental document. The codebase and documentation are already linked with the paper.

# 6 Benchmarking Algorithms on SustainDC

The purpose of SustainDC is to understand the benefits jointly optimizing the *Workload, Data Center* and *Battery Environments* for reducing the operating *CFP* of a DC. To investigate this idea, first, we consider the net operating *CFP* for instances where we only evaluate with a trained RL agent for one of the SustainDC environment while employing the baseline methods ($B_*$) for the other environments. The baseline method ($B_{LS}$) for the *Workload Environment* ($Env_{LS}$) simply implies that no workload is moved over the horizon, which is the current standard in the majority of the DCs. For the *Data Center Environment* ($Env_{DC}$), we consider the industry standard ASHRAE Guideline 36 as the baseline ($B_{DC}$) Zhang et al. [2022]. For the *Battery Environment* ($Env_{BAT}$), we consider a real time adaptation of the method in Acun et al. [2023b] as the baseline ($B_{BAT}$) (i.e.: we reduced the optimization horizon in the original paper from 24 hours to 3 hours). In future work, we shall include further baseline comparisons using MPCs and other non-ML control algorithms. Next, we perform ablations on the collaborative reward parameter $a$. Finally, then benchmark variants of the multi-agent RL approach. This includes multiagent PPO Schulman et al. [2017] with independent critic associated with each actor network (IPPO) de Witt et al. [2020], a centralized critic that has access to states and actions from other MDPs (MAPPO) Yu et al. [2022]. Given the heterogeneous nature of the action and observation spaces for SustainDC, we also benchmark several Heterogeneous multi-agent RL (HARL) Zhong et al. [2024] approaches including HAPPO (Heterogeneous Agent PPO), HAA2C (Heterogeneous Agent Advantage Actor Critic), HAD3QN (Heterogeneous Agent Dueling Double Deep Q Network) and HASAC (Heterogeneous Agent Soft Actor Critic). For our experiments, the MARL agents were trained on one location and evaluated for different locations.

In Figure 4, we compare the relative performance of different RL algorithms on a radar chart based on the evaluation metrics in Section 5. Since reporting absolute values may not convey much sense, we plot the relative differences in RL agent performances, which provides insights in to the *pros* and *cons* of each approach. (We will provide the absolute values for these benchmark experiments in the supplementary document in tabular format.) Hence, we normalize the absolute values of the metrics of interest for each benchmarking approach with respect to the average and standard deviation. We consider the lowest values towards the periphery of the radar chart and higher values towards the center. Hence, the larger the area of each approach on the radar chart, the better is the approach w.r.t the metrics of interest.

## 6.1 Single vs multi-agent Benchmarks

Figure 4a compares the relative performance of a single RL agent vs multi-agent RL benchmarks to motivate the use of a MARL approach for sustainable DC operation. Among single RL agent approaches, the workload manager RL agent (Experiment 1) and the battery agent (Experiment 3) performs similarly reducing water usage. The standalone DC RL agent (Experiment 2) performs well overall on energy and *CFP* reduction. (For Experiments 1 and 3, the performance on the Lowest Task Queue metric should be ignored, since the baseline workload manager doesn't shift workload and by default has the lowest task queue). When we consider pairs of RL agents working simultaneously, the absence of cooling optimization agent (i.e. Experiment 5) leads to almost similar results as single RL agent implementations (i.e., Experiments 1 and 3), where we only use $A_{LS}$ or $A_{BAT}$ with other baseline agents. This shows that the RL based cooling optimizer improves overall performance compared to the rule-based Guideline 36 controller (i.e., Experiments 2 and 4). Finally, when we consider all three RL agents operating simultaneously without a shared critic (Experiment 7 using IPPO), they achieve better energy consumption, water usage and task queue and relatively similar *CFP* than other experiments. Hence, the combined performance across all three agents motivates the use of MARL.

## 6.2 Reward Ablation on $a$

Figure 4b, shows the relative differences in performance when considering collaborative reward components. We considered 2 values of $a$ at the extremes to indicate no collaboration ($a = 1.0$) and relying only on the rewards of other agents ($a = 0.1$). An intermediate value of $a = 0.8$ was chosen based on similar work on reward-based collaborative approach in Sarkar et al. [2023d, 2022]. The improvement in setting $a = 0.8$ shows that considering rewards from other agents can improve performance w.r.t. no collaboration ($a = 1.0$) especially in a partially observable MDP.

## 6.3 Multiagent Benchmarks

We evaluated and compared the relative performances of different MARL that includes PPO with independent actor critics (IPPO, $a = 0.8$), centralized critic PPO (MAPPO), heterogeneous multiagent PPO (HAPPO), HAA2C, HAD3QN and HASAC. Figures 4c, 4d, 4e, 4f show the relative performance of these approaches for data centers located in New York, Georgia, and California and Arizona. We observed a clear trend where PPO based shared actor critic methods (MAPPO, HAPPO) outperform their independent agent counterpart IPPO. On closer investigation of the actions, we found that while IPPO is able to reduce HVAC and IT energy, the battery agent is not able to schedule when to charge from the grid and discharge optimally to meet data center demand. Among MAPPO, HAPPO and HA2C, the HAPPO was consistently able to perform better (except Georgia). Among the off-policy methods (HAD3QN and HASAC), there is a significant variance in performance across different regions, with HASAC outperforming all others in Arizona. We have not been able to fully understand the overall reason for these variations which can be partially attributed to weather and carbon intensity related changes. We will continue to investigate this changes in future work.

## 7 Limitations

The absence of an oracle that already knows the best results possible for the different environments makes it difficult to quantify the threshold for performance compared to simpler environments. For

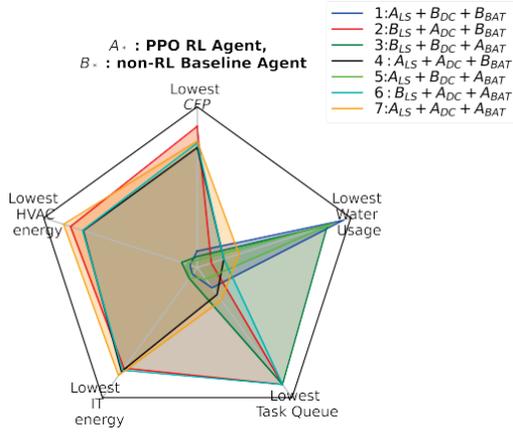

(a) Single RL agent, two RL agents and three RL agents
For single agents, PPO was used (Average result over
5 runs)

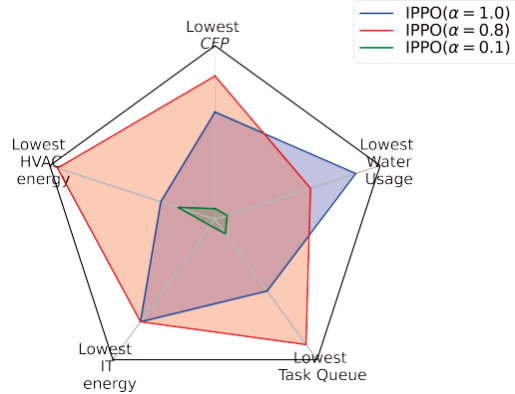

(b) IPPO with different values of collaborative reward
coefficient $\alpha$ (Average result over 12 runs)

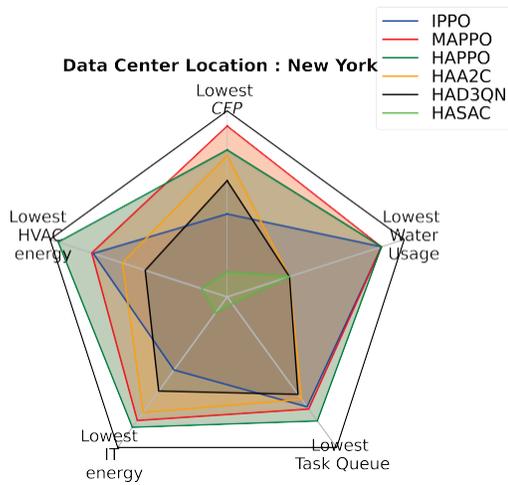

(c) Multiagent RL frameworks for a data center located
in New York (Average result over 5 runs)

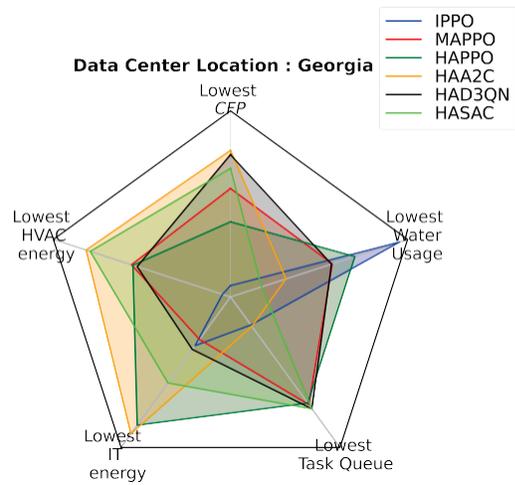

(d) Multiagent RL frameworks for a data center located
in Georgia (Average result over 5 runs)

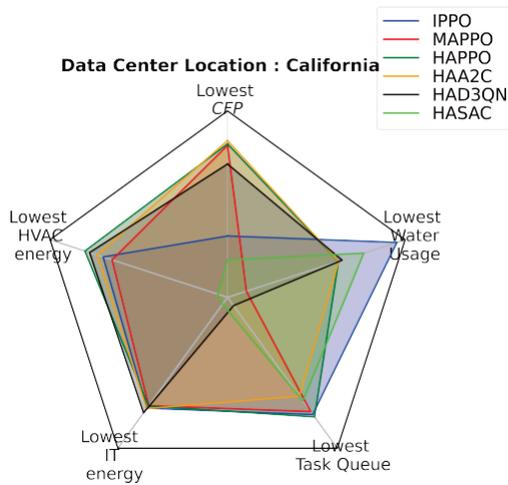

(e) Multiagent RL frameworks for a data center located
in California (Average result over 5 runs)

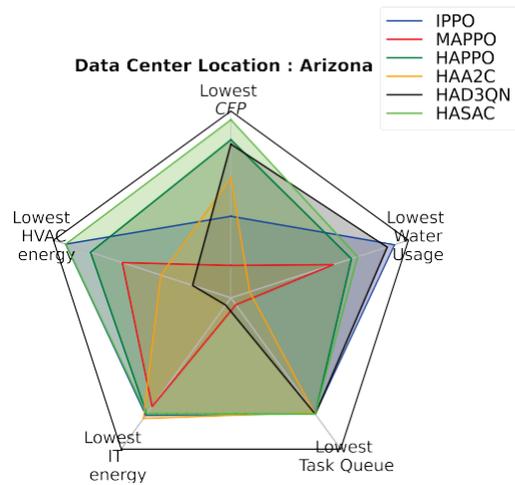

(f) Multiagent RL frameworks for a data center located
in Arizona (Average result over 5 runs)

Figure 4: Benchmarking RL Algorithms on the Sustain DC environment

computational speed in RL, we used reduced order models for certain components like pumps and cooling towers. We could not exhaustably tune the hyperparameters for all the networks.

## 8   Next Steps

We are planning to deploy the trained agents to real data centers and are working towards domain adaptation for deployment with safeguards. We will augment the codebase with these updates. In order to have a smooth integration with current systems where HVAC runs in isolation, we plan a phased deployment with recommendation to the data center operative followed by direct integration of the control agents with the HVAC system with safeguards. For real-world deployment, a trained model should be run on a production server using appropriate checkpoints within a containerized platform with necessary dependencies. Security measures must restrict the software to only read essential data, generate decision variables, and write them with limited access to secure memory for periodic reading by the data center's HVAC management system. To ensure robustness against communication loss, a backup mechanism for generating decision variables is essential.

## 9   Conclusion

In this paper, we introduced SustainDC, a MARL benchmarking environments developed completely in Python. It allows the user to solve problems related to sustainable, cost and energy efficient data center operations. SustainDC allows the customization of every aspect of the data center. It allows a great degree of customization w.r.t. the RL reward design, which is an important problem that we invite other researchers to collaborate on using SustainDC. We benchmark an extensive collection of single and multiagent RL algorithms on SustainDC across multiple geographical locations and compare their performance. This can provide researchers with the background to sustainably control data centers using reinforcement learning. In this regard, we are working on realizing some of the proposed methods with consortiums like ExaDigiT which focuses on HPC and Supercomputing along with other customers and industry collaborators. This can also be used to benchmark hierarchical RL algorithms given the complexity and constraints of this environment rooted in a real-world system. Furthermore, to plan for equipment changes and data center accessories, we will implement continual reinforcement learning to avoid out-of-distribution errors.

# A Models

## A.1 Workload Environment ($Env_{LS}$)

The Workload Environment ($Env_{LS}$) simulates the management and scheduling of data center (DC) workloads, allowing for dynamic adjustment of utilization to optimize energy consumption and carbon footprint. The environment is designed to evaluate the performance of reinforcement learning (RL) algorithms in rescheduling delayable workloads within the DC.

Let $B_t$ be the instantaneous DC workload trace at time $t$, with $X\%$ of the load being rescheduled up to $N$ simulation steps into the future. The goal of an RL agent ($Agent_{LS}$) is to observe the current time of day ($SC_t$), the current and forecast grid CI data ($CI_{t...t+L}$), and the amount of rescheduled workload left ($D_t$). Based on these observations, the agent decides an action $A_{ls,t}$ to reschedule the flexible component of $B_t$ to create a modified workload $\hat{B}_t$, thus minimizing the net $CFP = \sum_{t=0}^{N} CFP_t$ over $N$ steps. Here $CFP_t$ will be calculated based on the sum of the DC IT load due to $\hat{B}_t$, the corresponding HVAC cooling load, and the charging and discharging of the battery at every time step.

### A.1.1 Actions ($A_{LS}$)

The action space for $Agent_{LS}$ includes three discrete actions:

- *Action 0: Decrease Utilization* - This action attempts to defer the flexible portion of the current workload ($B_{nonflex}$) to a later time. The non-flexible ($B_{flex}$) workload is processed immediately, while the flexible workload is added to a queue for future execution.
- *Action 1: Do Nothing* - This action processes both the flexible ($B_{flex}$) and non-flexible ($B_{nonflex}$) portions of the current workload immediately without any deferral.
- *Action 2: Increase Utilization* - This action attempts to increase the current utilization by processing tasks from the queue, if available, in addition to the current workload.

### A.1.2 Observations ($S_{LS}$)

The state space observed by the RL agent consists of several features, including:

- **Time of Day** - Represented using sine and cosine transformations of the hour of the day to capture cyclical patterns.
- **Day of the Year** - Represented using sine and cosine transformations to capture seasonal variations.
- **Current Workload** - The current workload level, which includes both flexible and non-flexible components.
- **Queue Status** - The length of the task queue, normalized by the maximum queue length.
- **Grid Carbon Intensity (CI)** - Current and forecasted CI values, capturing the environmental impact of electricity consumption.
- **Battery State of Charge (SoC)** - The current state of charge of the battery, if available.

The observation space is a combination of these features, providing the agent with a comprehensive view of the current state of the environment.

### A.1.3 Mathematical Model

**Workload Breakdown** Let $B_t$ be the total workload at time $t$. This workload is divided into flexible ($B_{flex,t}$) and non-flexible ($B_{nonflex,t}$) components:

$$B_t = B_{flex,t} + B_{nonflex,t}$$

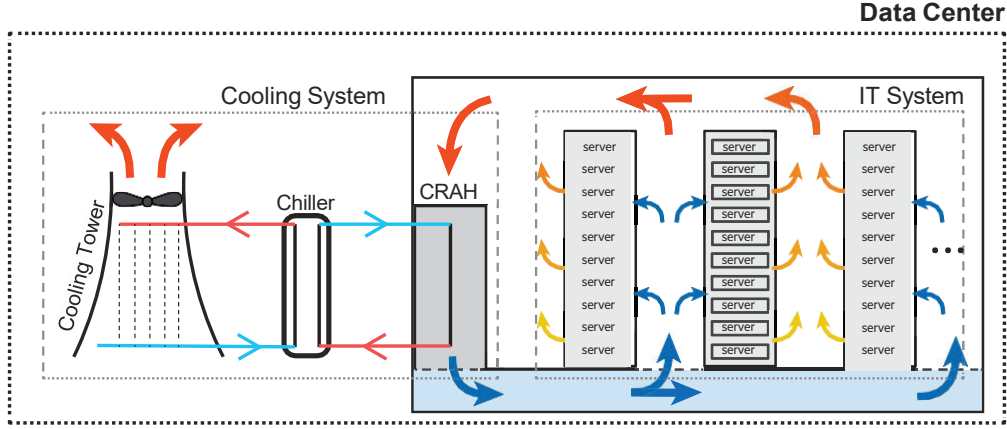

Figure 5: Illustration of the modeled data center, showing the IT section (cabinets and servers) and the Cooling section (Cooling Tower, chiller, and CRAH). The airflow path is also depicted, with cool air supplied through the raised floor and hot air returning via the ceiling. Note: We use CRAH and CRAC interchangeably in the text, but they both represent the same device (CRAH).

The flexible workload $B_{flex,t}$ is a fraction of the total workload:

$$B_{flex,t} = a \cdot B_t, \quad 0 < a < 1$$

where $a$ is the flexible workload ratio.

**Actions and Workload Management**  Depending on the action $A_{ls,t}$ chosen by the RL agent, the workload is managed as follows:

1. *Action 0: Decrease Utilization (Queue Flexible Workload)*

$$\hat{B}_t = B_{nonflex,t}$$

The flexible workload $B_{flex,t}$ is added to a task queue $Q_t$ for future execution:

$$Q_{t+1} = Q_t + B_{flex,t}$$

2. *Action 1: Do Nothing*

$$\hat{B}_t = B_t = B_{nonflex,t} + B_{flex,t}$$

There is no change in the task queue:

$$Q_{t+1} = Q_t$$

3. *Action 2: Increase Utilization (Process Queue)*

$$\hat{B}_t = B_t + \min(Q_t, C_{max} - B_t)$$

where $C_{max}$ is the maximum processing capacity. The processed tasks are removed from the task queue:

$$Q_{t+1} = Q_t - \min(Q_t, C_{max} - B_t)$$

### A.2 Data Center Environment ($Env_{DC}$)

The Data Center Environment ($Env_{DC}$) simulates the IT and HVAC operations within a DC, enabling the evaluation of RL algorithms aimed at optimizing cooling setpoints to reduce energy consumption and carbon footprint.

The data center modeled is illustrated in Figure 5. The IT section includes the cabinets and servers, while the Cooling section comprises a Cooling Tower, a chiller, and the Computer Room Air Handler (CRAH). The setup also features a raised floor system that channels cool air from the CRAH to the cabinets. The hot air exits the cabinets and returns to the CRAH via the ceiling.

### A.2.1 Data Center IT Model

Let $\hat{B}_t$ be the net DC workload at time instant $t$ obtained from the Workload Manager. The spatial temperature difference, $\mathbf{\Delta T}_{supply}$, given the DC configuration, is obtained from Computational Fluid Dynamics (CFD). For a given rack, the inlet temperature $T_{inlet,i}$ at $CPU_i$ is computed as:

$$T_{inlet,i,t} = \mathbf{\Delta T}_{supply,i} + T_{CRACsupply,t}$$

where $T_{CRACsupply,t}$ is the CRAC unit supply air temperature. This value is chosen by the RL agent $A_{DC}$.

Next, the CPU power curve $f_{cpu}(inlet\_temp,\ cpu\_load)$ and IT Fan power curve $f_{itfan}(inlet\_temp,\ cpu\_load)$ are implemented as linear equations based on Sun et al. [2021]. Given a server inlet temperature of $T_{inlet,i,t}$ and a processing amount of $\hat{B}_t$ performed by $CPU_i$, the total rack power consumption for rack $k$ across all CPUs from $i = 1$ to $K$, and the total DC Power IT Consumption can be calculated as follows:

$$P_{CPU,t} = \sum_i f_{cpu}(T_{inlet,i,t}, \tilde{B}_t)$$

$$P_{IT\ Fan,t} = \sum_i f_{itfan}(T_{inlet,i,t}, \tilde{B}_t)$$

$$P_{rack,k,t} = P_{CPU,t} + P_{IT\ Fan,t}$$

$$P_{datacenter,t} = \sum_k P_{rack,k,t}$$

### A.2.2 HVAC Cooling Model

Based on the DC IT Load $P_{datacenter,t}$, the IT fan airflow rate, $V_{sfan}$, air thermal capacity $C_{air}$, and air density, $\rho_{air}$, the rack outlet temperature $T_{outlet,i,t}$ is estimated from Sun et al. [2021] using:

$$T_{outlet,i,t} = T_{inlet,i,t} + \frac{P_{rack,k,t}}{C_{air} \cdot \rho_{air} \cdot V_{sfan}}$$

In conjunction with the return temperature gradient information $\mathbf{\Delta T}_{return}$ estimated from CFDs, the final CRAC return temperature is obtained as:

$$T_{CRACreturn,t} = \mathrm{avg}(\mathbf{\Delta T}_{return,i} + T_{outlet,i,t})$$

We assume a fixed-speed CRAC Fan unit for circulating air through the IT Room. Hence, the total HVAC cooling load for a given CRAC setpoint $T_{CRACsupply,t}$, return temperature $T_{CRACreturn,t}$, and the mass flow rate $m_{crac,fan}$ is calculated as:

$$P_{cool,t} = m_{crac,fan} \cdot C_{air} \cdot (T_{CRACreturn,t} - T_{CRACsupply,t})$$

To perform $P_{cool,t}$, the amount of cooling, the net chiller load for a chiller with Coefficient of Performance ($COP$) may be estimated as:

$$P_{chiller,t} = P_{cool,t} \left\{ 1 + \frac{1}{COP} \right\}$$

Next, this cooling load is passed on to the cooling tower. Assuming a cooling tower delta as a function of temperature $f_{ct\_delta}(t_{db})$, Breen et al. [2010] the required cooling tower air flow rate is calculated as:

$$V_{ct,air,t} = \frac{P_{chiller,t}}{C_{air} \cdot \rho_{air} \cdot f_{ct\_delta}(t_{db})}$$

Finally, the Cooling Tower Load at a flow rate of $V_{ct,air,t}$ is calculated with respect to a reference air flow rate $V_{ct,air,REF}$ and power consumption $P_{ct,REF}$ from the configuration object:

$$P_{CT,t} = P_{ct,REF} \left\{ \frac{V_{ct,air,t}}{V_{ct,air,REF}} \right\}^{-3}$$

Thus, the total HVAC load includes the cooling tower and chiller loads:

$$P_{HVAC,t} = P_{CT,t} + P_{chiller,t}$$

Based on these power values, the IT and HVAC Cooling energy consumptions can be represented as:

$$E_{hvac,t} = P_{HVAC,t} \times \text{step size} \tag{9}$$

$$E_{it,t} = P_{datacenter,t} \times \text{step size} \tag{10}$$

### A.2.3   Actions ($A_{DC}$)

The action space for $Agent_{DC}$ consists of discrete actions representing the adjustment of the CRAC unit's supply air temperature, limited to a range between 16°C to 23°C:

- *Action 0: Decrease Temperature* - The agent decreases the CRAC supply air temperature, enhancing cooling performance but increasing energy consumption.
- *Action 1: Maintain Temperature* - The agent maintains the current CRAC supply air temperature.
- *Action 2: Increase Temperature* - The agent increases the CRAC supply air temperature, which can reduce cooling energy consumption but may increase the IT equipment temperature.

### A.2.4   Observations ($S_{DC}$)

The state space observed by the RL agent consists of several features, including:

- **Time of Day** - Represented using sine and cosine transformations of the hour of the day to capture cyclical patterns.
- **Day of the Year** - Represented using sine and cosine transformations to capture seasonal variations.
- **Ambient Weather** - Includes current temperature and other relevant weather conditions.
- **IT Room Temperature** - Average temperature in the IT room.
- **Energy Consumption** - Previous step cooling and IT energy consumptions.
- **Grid Carbon Intensity (CI)** - Current and forecasted CI values.

The observation space provides a comprehensive view of the current state of the environment to the agent.

### A.2.5   Chiller Sizing

The chiller power consumption is calculated based on the load and operating conditions using the following method:

$$P_{chiller,t} = \text{calculate\_chiller\_power}(max\_cooling\_cap, load, ambient\_temp)$$

**Calculation of Average CRAC Return Temperature**

$$T_{CRACreturn,t} = \text{avg}(\mathbf{\Delta T}_{return,i} + T_{outlet,i,t})$$

**Calculation of HVAC Power**

$$P_{cool,t} = m_{crac,fan} \cdot C_{air} \cdot (T_{CRACreturn,t} - T_{CRACsupply,t})$$

$$P_{chiller,t} = P_{cool,t} \left\{ 1 + \frac{1}{COP} \right\}$$

$$V_{ct,air,t} = \frac{P_{chiller,t}}{C_{air} \cdot \rho_{air} \cdot f_{ct\_delta}(t_{dp})}$$

$$P_{CT,t} = P_{ct,REF} \left\{ \frac{V_{ct,air,t}}{V_{ct,air,REF}} \right\}^{-3}$$

$$P_{HVAC,t} = P_{CT,t} + P_{chiller,t}$$

### A.2.6 Water Consumption Model

The water usage for the cooling tower is estimated using a model based on research findings from several key sources. The model accounts for the water loss due to evaporation, drift, and blowdown. The primary references used to develop this model include Sharma et al. [2009], Shublaq and Sleiti [2020], and guidelines from SPX Cooling Technologies SPX Cooling Technologies [2023].

The water usage model is formulated as follows:

1. **Range Temperature Calculation**: The difference between the hot water temperature entering the cooling tower and the cold water temperature leaving the cooling tower:

$$\text{range\_temp} = \text{hot\_water\_temp} - \text{cold\_water\_temp}$$

where hot_water_temp is the $T_{CRACreturn,t}$, and cold_water_temp is the current CRAC setpoint $T_{CRACsupply,t}$.

2. **Normalized Water Usage**: The baseline water usage per unit time, adjusted for the wet bulb temperature of the ambient air. This accounts for the environmental conditions affecting the cooling tower's efficiency:

$$\text{norm\_water\_usage} = 0.044 \cdot \text{wet\_bulb\_temp} + (0.35 \cdot \text{range\_temp} + 0.1)$$

3. **Total Water Usage**: The normalized water usage is adjusted to ensure non-negativity and further adjusted for drift losses, which are a small percentage of the total water circulated in the cooling tower:

$$\text{water\_usage} = \max(0, \text{norm\_water\_usage}) + \text{norm\_water\_usage} \cdot \text{drift\_rate}$$

4. **Water Usage Conversion**: The total water usage is converted to liters per simulation timestep interval for ease of reporting and consistency with other metrics. Given that we use $N$ timesteps per hour in our simulations, the conversion is as follows:

$$\text{water\_usage\_liters\_per\_timestep} = \left\{ \frac{\text{water\_usage} \cdot 1000}{N} \right.^{\text{L}}$$

This model incorporates both theoretical and empirical insights, providing a comprehensive estimation of the water consumption in a data center's cooling tower. By considering the specific operational parameters and environmental conditions, it ensures accurate and reliable water usage calculations, critical for sustainable data center management.

## A.3 Battery Environment ($Env_{BAT}$)

The Battery Environment ($Env_{Bat}$) simulates the battery banks operations within the DC, enabling the evaluation of RL algorithms aimed at optimizing auxiliary battery usage to reduce energy costs and carbon footprint. This environment is a modified version of the battery model from Acun et al. [2023a].

### A.3.1 Battery Model

The battery model represents the energy storage system, considering its capacity, charging and discharging efficiency, and rate limits. The battery state of charge (SoC) evolves based on the actions taken by the RL agent.

Let $E_{bat,t}$ be the energy stored in the battery at time $t$. The battery can perform three actions: charge, discharge, or remain idle. The maximum battery capacity is $C_{max}$, and the current state of charge is $E_{bat,t}$.

### A.3.2 Actions ($A_{Bat}$)

The action space for $Agent_{Bat}$ includes three discrete actions:

- *Action 0: Charge* - The battery is charged at a rate of $r_{charge}$, consuming $E_{bat,t}$ Wh of energy.

- *Action 1: Idle* - The battery do not consume energy.

- *Action 2: Discharge* - The battery discharges energy at a rate of $r_{discharge}$, supplying $E_{bat,t}$ Wh of energy.

### A.3.3 Observations ($S_{Bat}$)

The state space observed by the RL agent consists of several features, including:

- **Data Center Load** - The current power consumption of the data center.

- **Battery SoC** - The current state of charge of the battery.

- **Grid Carbon Intensity (CI)** - Current and forecasted CI values.

- **Time of Day and Year** - Represented using sine and cosine transformations to capture cyclical patterns.

The observation space is a combination of these features, providing the agent with a comprehensive view of the current state of the environment.

### A.3.4 Mathematical Model

**Battery Charging and Discharging** The energy stored in the battery evolves based on the action taken:

$$E_{bat,t} = \begin{cases} r_{charge} \cdot \eta_{charge} \cdot \Delta t & \text{if charging} \\ 0 & \text{if idle} \\ r_{discharge} \cdot \eta_{discharge} \cdot \Delta t & \text{if discharging} \end{cases}$$

where $r_{charge}$ and $r_{discharge}$ are the rates of charging and discharging the battery, respectively. These rates determine the amount of energy added to or removed from the battery within a time step $\Delta t$.

**Charging Rate ($r_{charge}$)** The charging rate $r_{charge}$ is the rate at which energy is added to the battery during the charging process. It is defined as:

$$r_{charge} = \min\left\{ \frac{C_{max} - E_{bat,t}}{\eta_{charge} \cdot \Delta t}, P_{charge,max} \right\}$$

where $P_{charge,max}$ is the maximum allowable charging power. This rate ensures that the battery does not exceed its maximum capacity $C_{max}$ and that charging occurs efficiently.

**Discharging Rate ($r_{discharge}$)** The discharging rate $r_{discharge}$ is the rate at which energy is drawn from the battery during the discharging process. It is defined as:

$$r_{discharge} = \min\left\{ \frac{E_{bat,t}}{\eta_{discharge} \cdot \Delta t}, P_{discharge,max} \right\}$$

where $P_{discharge,max}$ is the maximum allowable discharging power. This rate ensures that the battery does not discharge below zero and that discharging occurs efficiently.

**Energy Constraints** The state of charge is bounded by the battery capacity:

$$0 \leq E_{bat,t} \leq C_{max}$$

**Battery Power Constraints** The maximum power that the battery can charge or discharge is limited by:

$$P_{charge,max} = u \cdot P_{charge} + v$$

$$P_{discharge,max} = u \cdot P_{discharge} + v$$

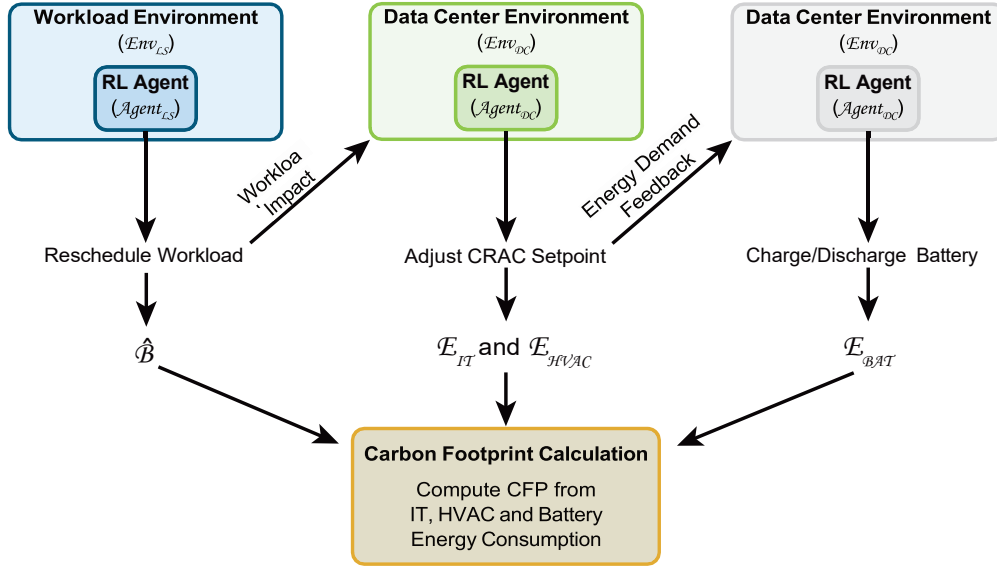

Figure 6: Interconnection of environments and agent actions. The figure shows how the Workload Environment ($Env_{LS}$) interacts with the Data Center Environment ($Env_{DC}$) by rescheduling workloads, and how the Data Center Environment impacts the Battery Environment ($Env_{BAT}$) through energy demands. Each agent observes the state of its respective environment and takes actions to optimize operations, with the overall goal of minimizing the carbon footprint (CFP) through coordinated efforts.

**Simple Reward Calculation** The goal of the three agents ($Agent_{LS}$, $Agent_{DC}$, and $Agent_{BAT}$) is to minimize the cumulative carbon footprint (CFP) over a given horizon $N$. The CFP at each time step $t$ is computed as:

$$CFP_t = (E_{it,t} + E_{hvac,t} + E_{bat,t}) \cdot CI_t$$

where:

- $E_{it,t}$: Energy consumption by IT equipment due to $\hat{B}_t$

- $E_{hvac,t}$: Energy consumption by HVAC systems

- $E_{bat,t}$: Energy contribution from the battery (positive when discharging, negative when charging)

- $CI_t$: Grid carbon intensity at time $t$

The total reward is then:

$$R = -\sum_{t=0}^{N} CFP_t$$

The reward could have other terms that may consider queue length, water usage, average task delay, etc.

## A.4 Interconnection of Environments and Agent Actions

Figure 6 illustrates the interconnection of the different environments ($Env_{LS}$, $Env_{DC}$, and $Env_{BAT}$) and the actions of their respective RL agents. This diagram highlights how the decisions made by each agent impact the overall DC operations and contribute to the optimization of energy consumption and carbon footprint.

In the **Workload Environment** ($Env_{LS}$), the RL agent ($Agent_{LS}$) reschedules flexible workloads to optimize utilization. This action will influence the IT load, which directly impacts the **Data Center Environment** ($Env_{DC}$). The RL agent ($Agent_{DC}$) in the data center environment adjusts the CRAC setpoints to optimize cooling and IT operations, thus affecting the HVAC cooling load and overall energy consumption.

The **Battery Environment (*Env_BAT*)** is influenced by the energy demands of the data center environment. The RL agent (*Agent_BAT*) manages the charging and discharging of the battery to optimize energy usage and reduce the carbon footprint. The interconnections between these environments ensure that the agents work together to minimize the cumulative CFP by considering the energy consumption of IT, HVAC, and battery systems.

By observing the current state and forecast data, each agent makes informed decisions that contribute to the overall sustainability and efficiency of the data center operations. This coordinated approach leverages the strengths of each environment to achieve significant reductions in energy consumption and carbon emissions.

# B  Customization of *dc_config.json*

The customization of the DC is done through the dc_config.json file located in the utils folder. This file allows users to specify every aspect of the DC environment design. We show here a part of the configuration file to indicate the different configurable elements inside SustainDC. Additional elements can be added to this config either under an existing section or a new section, and utils/dc_config_reader.py will automatically import the new configurations. Inside the "data_center_configuration" SustainDC allows the user to configure the dimensions of the data center arrangement, the compiled CFD supply and approach temperature delta values and the maximum allowable CPUs per rack. There is an extensive set of parameters that can be configured under the "hvac_configuration" section including physical constants, parameters of the computer room air-conditioning unit (CRAC), chiller, pumps and cooling towers. The "server_characteristics" block allows the user to specify the properties of individual servers in the data center, including their idle power, full load fan frequency and power.

```
{
"data_center_configuration"  :
{
    "NUM_ROWS" : 4,
    "NUM_RACKS_PER_ROW" : 5,
    "RACK_SUPPLY_APPROACH_TEMP_LIST" : [
                            5.3, 5.3, 5.3, 5.3,5.3,
                            5.0, 5.0, 5.0, 5.0,5.0,
                            5.0, 5.0, 5.0, 5.0,5.0,
                            5.3, 5.3, 5.3, 5.3, 5.3
                            ],
    "RACK_RETURN_APPROACH_TEMP_LIST" : [
                            -3.7, -3.7, -3.7, -3.7, -3.7,
                            -2.5, -2.5, -2.5, -2.5, -2.5,
                            -2.5, -2.5, -2.5, -2.5, -2.5,
                            -3.7, -3.7, -3.7, -3.7, -3.7
                            ],
    "CPUS_PER_RACK" : 200
},
"hvac_configuration"  :
{
    "C_AIR" : 1006,
    "RHO_AIR" : 1.225,
    "CRAC_SUPPLY_AIR_FLOW_RATE_pu" : 0.00005663,
    "CRAC_REFRENCE_AIR_FLOW_RATE_pu" : 0.00009438,
    "CRAC_FAN_REF_P"  : 150,
    "CHILLER_COP_BASE"  : 5.0,
    "CHILLER_COP_K"  : 0.1,
    "CHILLER_COP_T_NOMINAL" : 25.0,
    "CT_FAN_REF_P"  : 1000,
    "CT_REFRENCE_AIR_FLOW_RATE" : 2.8315,
```

```
            "CW_PRESSURE_DROP" : 300000,
            "CW_WATER_FLOW_RATE" : 0.0011,
            "CW_PUMP_EFFICIENCY" : 0.87,
            "CT_PRESSURE_DROP" : 300000,
            "CT_WATER_FLOW_RATE": 0.0011,
            "CT_PUMP_EFFICIENCY" : 0.87
        },
        "server_characteristics" :
        {
            "CPU_POWER_RATIO_LB" : [0.01, 1.00],
            "CPU_POWER_RATIO_UB" : [0.03, 1.02],
            "IT_FAN_AIRFLOW_RATIO_LB" : [0.01, 0.225],
            "IT_FAN_AIRFLOW_RATIO_UB" : [0.225, 1.0],
            "IT_FAN_FULL_LOAD_V" : 0.051,
            "ITFAN_REF_V_RATIO" : 1.0,
            "ITFAN_REF_P" : 10.0,
            "INLET_TEMP_RANGE" : [16, 28],
            "DEFAULT_SERVER_POWER_CHARACTERISTICS":[[170,  20],
                                                   [120, 10],
                                                   [130, 10],
                                                   [130, 10],
                                                   [130, 10],
                                                   [130, 10],
                                                   [130, 10],
                                                   [130, 10],
                                                   [130, 10],
                                                   [130, 10],
                                                   [130, 10],
                                                   [130, 10],
                                                   [170, 10],
                                                   [130, 10],
                                                   [130, 10],
                                                   [110, 10],
                                                   [170, 10],
                                                   [170, 10],
                                                   [170, 10]],
            "HP_PROLIANT" : [110,170]
        }
}
```

## C  Performance of RL agents on Evaluation Metrics

In this section, we provide the numerical results we obtained from the main paper. The results are shown in Tables 2 (advantage of multiagent vs single agent), 3 (effects of reward sharing across agents), 4, 5, 6 and 7 (ablation across geographical locations with different weather, grid carbon intensity and server load pattern). We observed that there is not a single algorithm that works well across different metrics and geographical locations, and this is visually appreciated in the main paper.

## D  Agents/Env behavior

### D.1  Battery

The battery environment demonstrates how the battery's state of charge (SoC) and actions evolve over time under random behaviors. These figures illustrate two different examples generated using distinct random seeds.

Table 2: Performance with respect to evaluation metrics on single and multiple RL agent baselines. $A_* : RL\ agent\ B_* : non-RL\ baseline\ agent$

| Evaluation Metric → Algorithm ↓ | $CFP$ (kgCO2) | HVAC Energy (kwh) | IT Energy (kwh) | Task Queue | Water Usage (litre) |
|---|---|---|---|---|---|
| 1: $A_{LS} + B_{DC} + B_{BAT}$ | 167.61 | 391.6 | 1033.8 | 0.52 | 10433.46 |
| 2: $B_{LS} + A_{DC} + B_{BAT}$ | 153.56 | 372.9 | 944.5 | 0.0 | 10930.77 |
| 3: $B_{LS} + B_{DC} + A_{BAT}$ | 168.22 | 390.3 | 1029.8 | 0.0 | 10493.95 |
| 4: $A_{LS} + A_{DC} + B_{BAT}$ | 155.97 | 374.9 | 941.3 | 0.48 | 10883.73 |
| 5: $A_{LS} + B_{DC} + A_{BAT}$ | 168.64 | 391.1 | 1030.9 | 0.56 | 10470.43 |
| 6: $B_{LS} + A_{DC} + A_{BAT}$ | 155.44 | 374.8 | 942.5 | 0 | 10883.73 |
| 7: $A_{LS} + A_{DC} + A_{BAT}$ | 155.23 | 371.8 | 937.4 | 0.45 | 10826.61 |

Table 3: IPPO evaluated on SustainDC with different values of collaborative reward coefficient $a$ (Average result over 12 runs)

| Evaluation Metric → Algorithm ↓ | $CFP$ (kgCO2) | HVAC Energy (kwh) | IT Energy (kwh) | Task Queue | Water Usage (litre) |
|---|---|---|---|---|---|
| IPPO($a = 1.0$) | 176.3 | 415.2 | 932.8 | 12.5 | 445.6 |
| IPPO($a = 0.8$) | 176.2 | 414.6 | 932.8 | 9.5 | 445.8 |
| IPPO($a = 0.1$) | 176.4 | 415.3 | 932.9 | 15.7 | 446.2 |

Table 4: Multiagent RL framework evaluated on SustainDC for a data center located in New York (Average result over 5 runs)

| Evaluation Metric → Algorithm ↓ | $CFP$ (kgCO2) | HVAC Energy (kwh) | IT Energy (kwh) | Task Queue | Water Usage (litre) |
|---|---|---|---|---|---|
| IPPO | 179.6 | 417.1 | 945.9 | 20.9 | 446.2 |
| MAPPO | 176.4 | 417.0 | 932.7 | 19.6 | 446.2 |
| HAPPO | 177.3 | 414.8 | 930.9 | 12.8 | 441.9 |
| HAA2C | 177.5 | 419.0 | 934.8 | 25.2 | 14977.1 |
| HAD3QN | 178.4 | 420.5 | 940.4 | 28.0 | 14950.9 |
| HASAC | 181.7 | 424.2 | 960.8 | 79.7 | 14842.4 |

Table 5: Multiagent RL framework evaluated on SustainDC for a data center located in Georgia (Average result over 5 runs)

| Evaluation Metric → Algorithm ↓ | $CFP$ (kgCO2) | HVAC Energy (kwh) | IT Energy (kwh) | Task Queue | Water Usage (litre) |
|---|---|---|---|---|---|
| IPPO | 265.4 | 376.7 | 935.4 | 6.8 | 31773.5 |
| MAPPO | 263.4 | 370.3 | 935.9 | 0.35 | 31949.9 |
| HAPPO | 264.1 | 370.4 | 929.0 | 0.47 | 31890.7 |
| HAA2C | 262.7 | 367.1 | 928.3 | 6.6 | 32071.5 |
| HAD3QN | 262.8 | 370.7 | 935.1 | 0.0 | 31952.2 |
| HASAC | 263.0 | 367.4 | 932.4 | 0.0 | 32135.7 |

Figure 7 shows the battery's SoC and the actions taken (Charge, Discharge, Idle) over simulated days for two different random behaviors.

Figure 8 compares the energy consumption with and without the battery over simulated days for two different random behaviors. This comparison illustrates the impact of battery usage on the overall energy consumption of the data center.

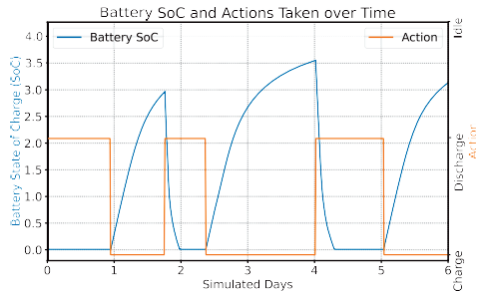

(a) Battery behavior example 1

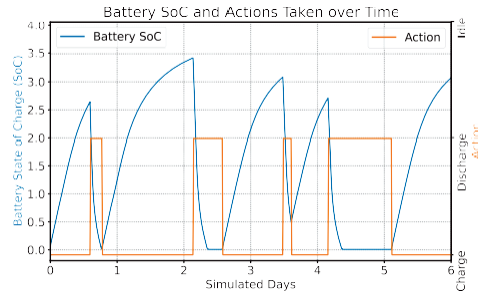

(b) Battery behavior example 2

Figure 7: Battery State of Charge (SoC) and actions taken over time under two different random behaviors. The actions are labeled as Charge, Discharge, and Idle.

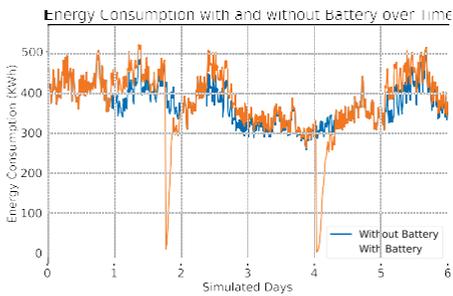

(a) Battery behavior example 1

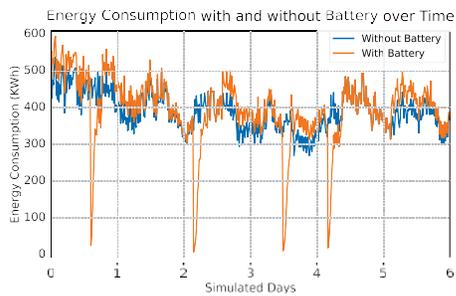

(b) Battery behavior example 2

Figure 8: Energy consumption with and without the battery over time under two different random behaviors. The comparison illustrates the effect of battery usage on overall energy consumption.

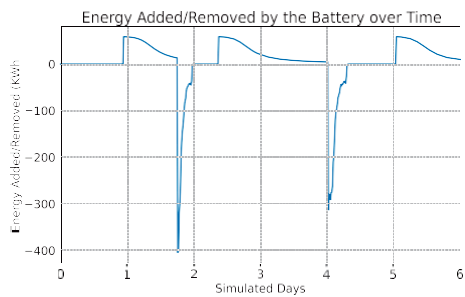

(a) Battery behavior example 1

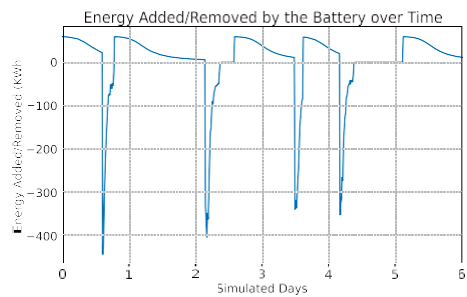

(b) Battery behavior example 2

Figure 9: Energy added to and removed from the battery over time under two different random behaviors. The figures show how the battery charges and discharges energy throughout the simulated period.

Table 6: Multiagent RL framework evaluated on SustainDC for a data center located in California (Average result over 5 runs)

| Evaluation Metric → Algorithm ↓ | $CFP$ (kgCO2) | HVAC Energy (kwh) | IT Energy (kwh) | Task Queue | Water Usage (litre) |
|---|---|---|---|---|---|
| IPPO | 170.0 | 384.3 | 933.8 | 12.9 | 28141.4 |
| MAPPO | 159.3 | 388.2 | 936.1 | 19.5 | 33289.3 |
| HAPPO | 159.1 | 376.3 | 935.8 | 74.9 | 30141.8 |
| HAA2C | 158.7 | 381.7 | 933.5 | 54.1 | 30135.4 |
| HAD3QN | 161.5 | 378.4 | 929.6 | 25.8 | 30017.4 |
| HASAC | 172.9 | 434.4 | 1027.0 | 43.8 | 29277.5 |

Table 7: Multiagent RL framework evaluated on SustainDC for a data center located in Arizona (Average result over 5 runs)

| Evaluation Metric → Algorithm ↓ | $CFP$ (kgCO2) | HVAC Energy (kwh) | IT Energy (kwh) | Task Queue | Water Usage (litre) |
|---|---|---|---|---|---|
| IPPO | 408.7 | 380.8 | 934.8 | 0.60 | 30251.6 |
| MAPPO | 410.8 | 383.3 | 947.5 | 502.4 | 31289.6 |
| HAPPO | 405.5 | 381.9 | 936.6 | 0.26 | 30983.7 |
| HAA2C | 407.1 | 385.0 | 929.9 | 7.54 | 32706.3 |
| HAD3QN | 405.6 | 386.4 | 1094.0 | 0.0051 | 30377.3 |
| HASAC | 404.6 | 380.8 | 936.7 | 0.54 | 30878.7 |

Figure 9 shows the energy added to and removed from the battery over simulated days for two different random behaviors. These figures demonstrate how the battery charges and discharges energy, providing insights into its operational patterns.

## E   External variables

### E.1   Workload

The *Workload* external variable in SustainDC represents the computational demand placed on the data center. Workload traces are provided in the form of FLOPs (floating-point operations) required by various jobs. By default, SustainDC includes a collection of open-source workload traces from *Alibaba* Alibaba Group [2017] and *Google* Google [2019] data centers. Users can customize this component by adding new workload traces to the *data/Workload* folder or specifying a path to existing traces in the *sustaindc_env.py* file under the workload_file configuration. Below is an example of modifying the workload configuration:

```
class EnvConfig(dict):

    DEFAULT_CONFIG = {
        "workload_file": "data/Workload/Alibaba_CPU_Data_Hourly_1.csv",
        ...
    }
```

The workload file should contain one year of data with an hourly periodicity (365*24=8760 rows). The file structure should have two columns, where the first column does not have a name, and the second column should be named cpu_load. Below is an example of the file structure:

```
,cpu_load
1,0.380
2,0.434
```

```
3,0.402
4,0.485
...
```

Figure 10 shows examples of different workload traces from Alibaba (v2017) and Google (v2011) data centers. Figure 11 provides a comparison between two workload traces of Alibaba (v2017) and Google (v2011).

## E.2 Weather

The *Weather* external variable in SustainDC captures the ambient environmental conditions impacting the data center's cooling requirements. By default, SustainDC includes weather data files in the .epw format from https://energyplus.net/weather for various locations where data centers are commonly situated. These locations include Arizona, California, Georgia, Illinois, New York, Texas, Virginia, and Washington. Users can customize this component by adding new weather files to the *data/Weather* folder or specifying a path to existing weather files in the *sustaindc_env.py* file under the weather_file configuration. Below is an example of modifying the weather configuration:

```
class EnvConfig(dict):

    DEFAULT_CONFIG = {
        'weather_file':   'data/Weather/USA_NY_New.York-Kennedy.epw',
        ...
    }
```

Each .epw file contains hourly data for various weather parameters, but for our purposes, we focus on the ambient temperature. Figure 12 shows the typical average ambient temperature across different locations over one year. Figure 13 provides a comparison of external temperatures across the different selected locations.

## E.3 Carbon Intensity

The *Carbon Intensity (CI)* external variable in SustainDC represents the carbon emissions associated with electricity consumption. By default, SustainDC includes CI data files for various locations: Arizona, California, Georgia, Illinois, New York, Texas, Virginia, and Washington. These files are located in the *data/CarbonIntensity* folder and are extracted from https://api.eia.gov/bulk/EBA.zip. Users can customize this component by adding new CI files to the *data/CarbonIntensity* folder or specifying a path to existing files in the *sustaindc_env.py* file under the cintensity_file configuration. Below is an example of modifying the CI configuration:

```
class EnvConfig(dict):

    DEFAULT_CONFIG = {
        'cintensity_file':   'data/CarbonIntensity/NY_NG_&_avgCI.csv',
        ...
    }
```

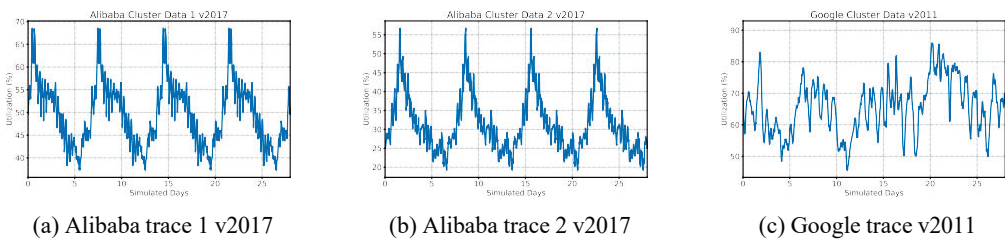

(a) Alibaba trace 1 v2017     (b) Alibaba trace 2 v2017     (c) Google trace v2011

Figure 10: Examples of different workload traces from Alibaba and Google data centers.

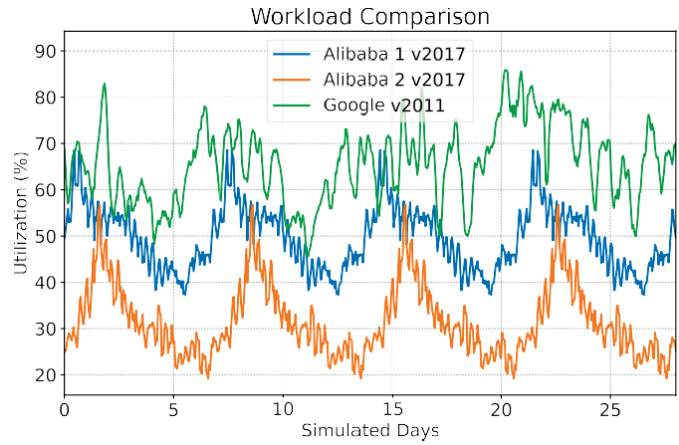

Figure 11: Comparison between two workload traces of Alibaba trace (2017) and Google (2011).

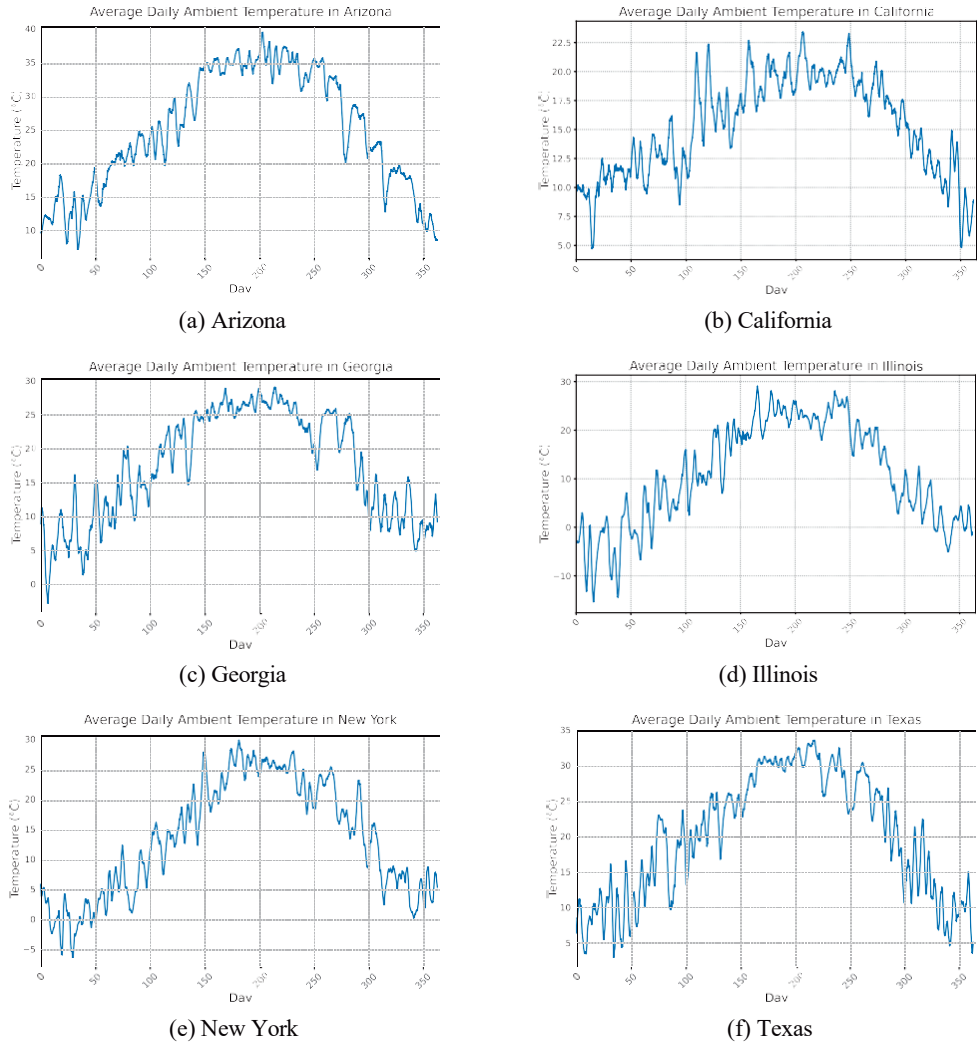

(a) Arizona

(b) California

(c) Georgia

(d) Illinois

(e) New York

(f) Texas

Figure 12: Typical average ambient temperature across different locations across one year.

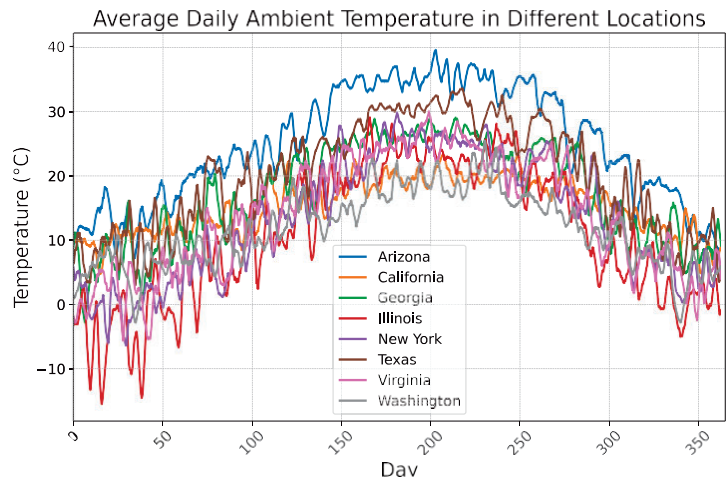

Figure 13: Comparison between external temperature of the different selected locations.

The CI file should contain one year of data with an hourly periodicity (365*24=8760 rows). The file structure should have the following columns: timestamp, WND, SUN, WAT, OIL, NG, COL, NUC, OTH, and avg_CI. WND, SUN, WAT, OIL, NG, COL, NUC, and OTH represent the energy sources contributing to the carbon intensity. These sources include wind, solar, water, oil, natural gas, coal, nuclear, and other types of energy, respectively. Below is an example of the file structure:

timestamp,WND,SUN,WAT,OIL,NG,COL,NUC,OTH,avg_CI
2022-01-01    00:00:00+00:00,1251,0,3209,0,15117,2365,4992,337,367.450
2022-01-01    01:00:00+00:00,1270,0,3022,0,15035,2013,4993,311,363.434
2022-01-01    02:00:00+00:00,1315,0,2636,0,14304,2129,4990,312,367.225
2022-01-01    03:00:00+00:00,1349,0,2325,0,13840,2334,4986,320,373.228
**...**

In Figure 14, the average daily carbon intensity for each selected location is shown, highlighting the variations in carbon emissions associated with electricity consumption across different regions.

In Figure 15, a comparison of carbon intensity across all the selected locations is presented, providing a comprehensive overview of how carbon emissions vary between these areas.

In Figure 16, we show the average daily carbon intensity against the average daily coefficient of variation (CV) for various locations. This figure highlights an important perspective on the variability and magnitude of carbon intensity values across different regions. Locations with a high CV indicate greater fluctuation in carbon intensity, offering more "room to play" for DRL agents to effectively reduce carbon emissions through dynamic actions. Additionally, locations with a high average carbon intensity value present greater opportunities for achieving significant carbon emission reductions. The selected locations are highlighted, while other U.S. locations are also plotted for comparison. Regions with both high CV and high average carbon intensity are identified as prime targets for DRL agents to maximize their impact on reducing carbon emissions.

In the table bellow (8) is the summarizing the selected locations, typical weather values, and carbon emissions characteristics:

Considering the data from Data Center Map, the U.S. states with the highest number of data centers are summarized in Table 9. The states with the most significant number of data centers tend to be Virginia, Texas, California, and New York. Virginia, especially, is a major hub due to its proximity to Washington D.C. and the abundance of fiber optic cable networks. Texas and California are also prominent due to their size, economic output, and significant tech industries. New York, particularly around New York City, hosts numerous data centers that serve the financial sector and other industries.

The selection of these locations is justified by their significant number of data centers, which emphasizes the potential impact of DRL agents in these regions. By targeting areas with both high data center density and favorable carbon intensity characteristics, DRL agents can maximize their effectiveness in reducing carbon emissions.

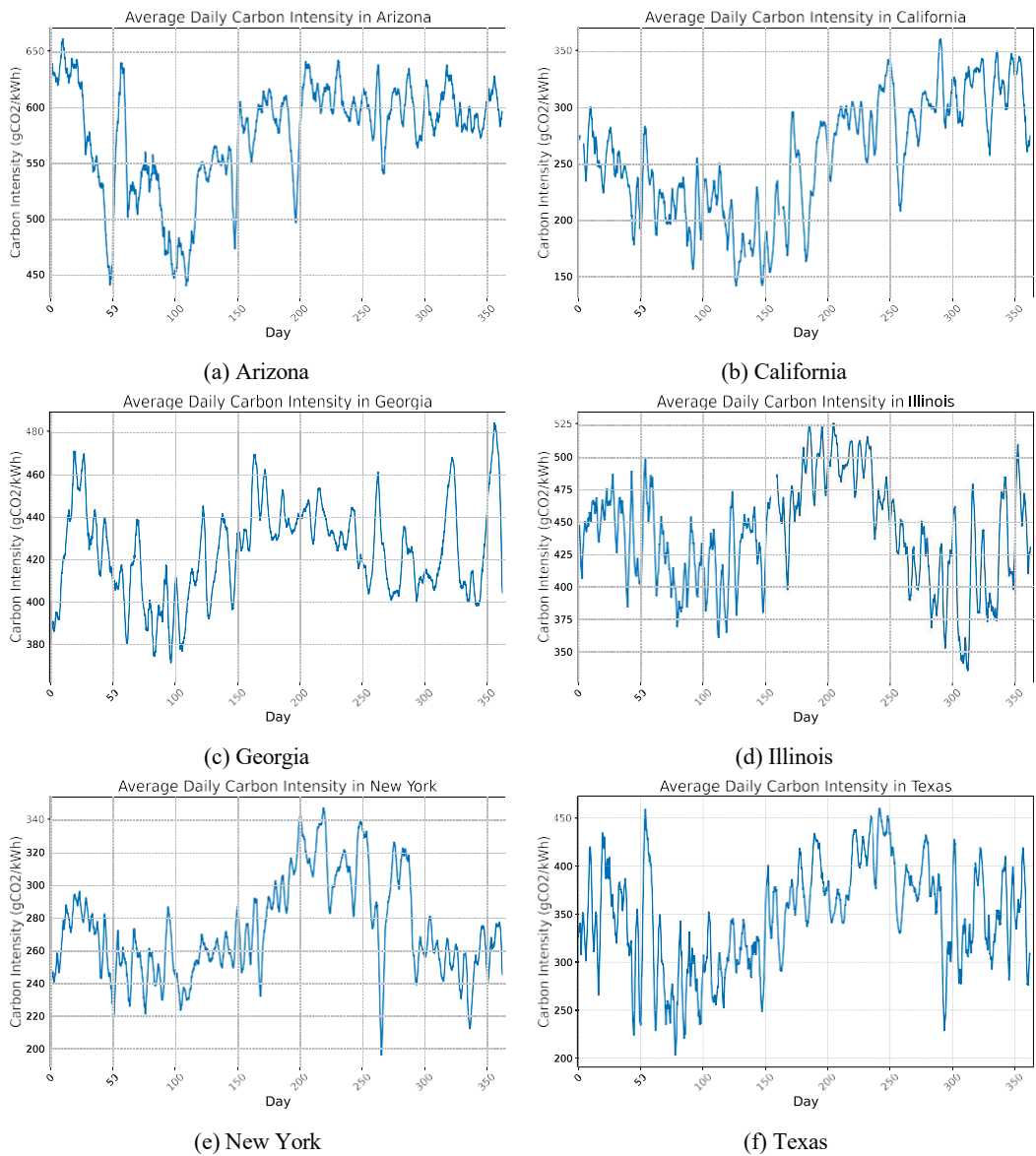

Figure 14: Typical average carbon intensity across different locations over one year.

| Location | Typical Weather | Carbon Emissions |
|---|---|---|
| Arizona | Hot, dry summers; mild winters | High avg CI, High variation |
| California | Mild, Mediterranean climate | Medium avg CI, Medium variation |
| Georgia | Hot, humid summers; mild winters | High avg CI, Medium variation |
| Illinois | Cold winters; hot, humid summers | High avg CI, Medium variation |
| New York | Cold winters; hot, humid summers | Medium avg CI, Medium variation |
| Texas | Hot summers; mild winters | Medium avg CI, Medium variation |
| Virginia | Mild climate, seasonal variations | Medium avg CI, Medium variation |
| Washington | Mild, temperate climate; wet winters | Low avg CI, Low variation |

Table 8: Summary of Selected Locations with Typical Weather and Carbon Emissions Characteristics

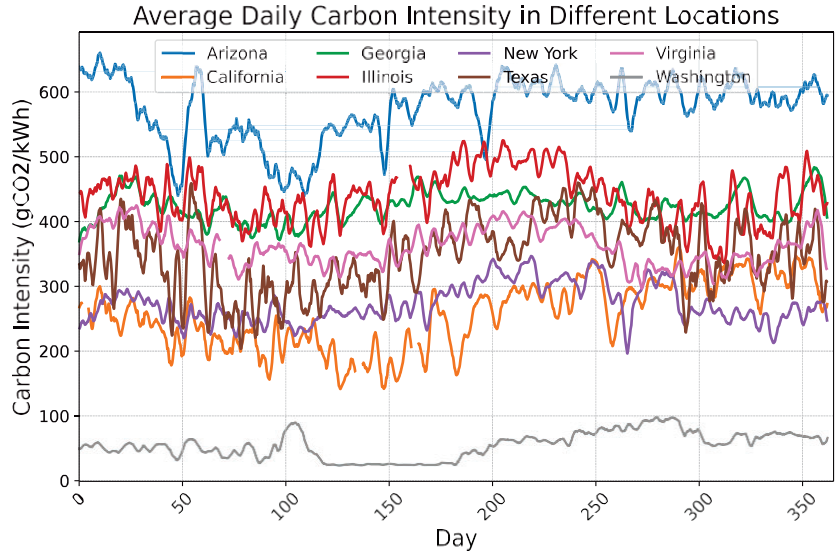

Figure 15: Comparison of carbon intensity across the different selected locations.

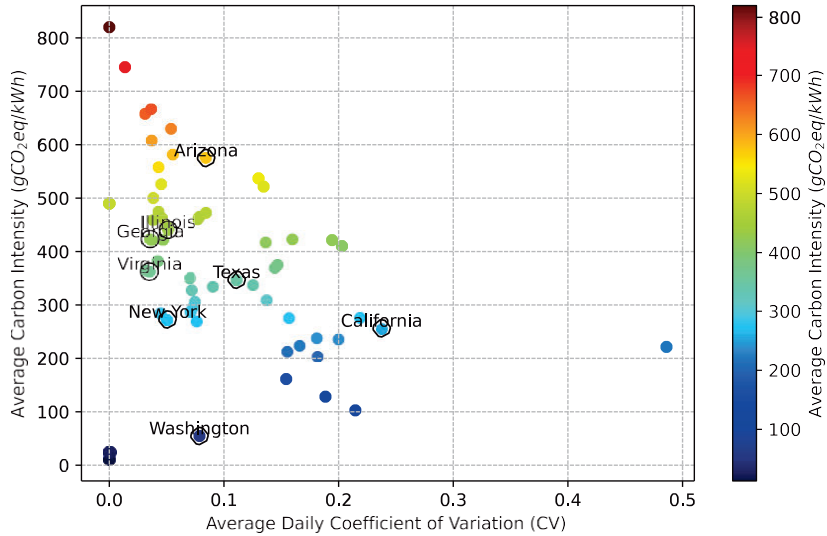

Figure 16: Average daily carbon intensity versus average daily coefficient of variation (CV) for the grid energy provided from US. Selected locations are remarked. High CV indicates more fluctuation, providing more opportunities for DRL agents to reduce carbon emissions. High average carbon intensity values offer greater potential gains for DRL agents.

| State | Data Centers |
|---|---|
| California | 254 |
| Virginia | 250 |
| Texas | 239 |
| New York | 128 |
| Illinois | 122 |
| Florida | 120 |
| Ohio | 98 |
| Washington | 84 |
| Georgia | 75 |
| New Jersey | 69 |

Table 9: Summary of U.S. States with the Most Data Centers (ref: Data Center Map)

## F Reward Evaluation and Customization

### F.1 Load Shifting Penalty ($LS_{Penalty}$)

The Load Shifting Penalty ($LS_{Penalty}$) is applied to the Load Shifting Agent ($Agent_{LS}$) in the Workload Environment ($Env_{LS}$) if it fails to reschedule flexible workloads within the same day. If $D_t$ (the amount of rescheduled workload left) is positive at the end of the day, $penalty\_tasks\_queue$ is assigned. Additionally, we included a function that progressively increases the penalty as the hour of the day approaches 24h. This means the penalty increases linearly from hour 23h to hour 24h.

Furthermore, there is a penalty for tasks that were dropped due to queue limits ($penalty\_dropped\_tasks$). This penalty is added to discourage the agent from dropping tasks and ensure that workloads are managed efficiently.

Therefore, the $LS_{Penalty}$ is composed of $penalty\_tasks\_queue$ and $penalty\_dropped\_tasks$. Related work in this area include Sarkar et al. [2024b,a, 2023g,f,e,c,b,a], Naug et al. [2023a,b], Athavale et al. [2024].

### F.2 Default Reward Function

The default reward function used in SustainDC for the Load Shifting Agent is implemented as follows:

```
def default_ls_reward(params: dict) -> float:
    """
    Calculate the reward value based on normalized load shifting
    and energy consumption.

    Parameters:
        params (dict): Dictionary containing parameters:
            - bat_total_energy_with_battery_KWh (float):
                Total energy consumption with battery.
            - norm_CI (float): Normalized carbon intensity.
            - bat_dcload_min (float): Minimum data center load.
            - bat_dcload_max (float): Maximum data center load.
            - ls_tasks_dropped (int): Number of tasks dropped due to queue limit.
            - ls_tasks_in_queue (int): Number of tasks currently in queue.
            - ls_current_hour (int): Current hour in the simulation.

    Returns:
        float: Calculated reward value.
    """
    # Energy part of the reward
    total_energy_with_battery = params['bat_total_energy_with_battery_KWh']
    norm_CI = params['norm_CI']
    dcload_min = params['bat_dcload_min']
    dcload_max = params['bat_dcload_max']

    # Calculate the reward associated with the energy consumption
    norm_net_dc_load = (total_energy_with_battery - dcload_min) /
                            (dcload_max - dcload_min)
    footprint = -1.0 * norm_CI * norm_net_dc_load
```

```
# Penalize the agent for each task that was dropped due to queue limit
penalty_per_dropped_task = -10    # Define the penalty value per dropped task
tasks_dropped = params['ls_tasks_dropped']
penalty_dropped_tasks = tasks_dropped * penalty_per_dropped_task

tasks_in_queue = params['ls_tasks_in_queue']
current_step = params['ls_current_hour']
penalty_tasks_queue = 0

if current_step % (24*4) >= (23*4):    # Penalty for queued tasks at the
                                                      end of the day
    factor_hour = (current_step % (24*4)) / 96       # min = 0.95833, max = 0.98953
    factor_hour = (factor_hour - 0.95833) / (0.98935 - 0.95833)
    penalty_tasks_queue = -1.0 * factor_hour * tasks_in_queue / 10          # Penalty
                                                      for each task left in the queue

LS_penalty = penalty_dropped_tasks + penalty_tasks_queue

reward = footprint + LS_penalty

return reward
```

## F.3  Customization of Reward Formulations

Users can choose to use any other reward formulation by defining custom reward functions inside *utils/reward_creator.py*. To create a custom reward function, you can define it as follows:

```
def custom_reward(params: dict) -> float:
    # Custom reward calculation logic
    pass
```

Replace the logic inside the custom_reward function with your custom reward logic.

For more examples of custom reward functions, users can check the file *utils/reward_creator.py*.

To use the custom reward function, you need to include it in the *utils/reward_creator.py* as follows:

```
# Other reward methods can be added here.

REWARD_METHOD_MAP = {
    'default_dc_reward' : default_dc_reward,
    'default_bat_reward': default_bat_reward,
    'default_ls_reward' : default_ls_reward,
    # Add custom reward methods here
    'custom_reward' : custom_reward,
}
```

Additionally, you need to specify the reward function in *harl/configs/envs_cfgs/dcrl.yaml*:

```
agents:
---
ls_reward: default_ls_reward
dc_reward: default_dc_reward
bat_reward: default_bat_reward
---
```

This flexibility ensures that SustainDC can be adapted to a wide range of research and operational needs in sustainable data center management.

# References

Bilge Acun, Benjamin Lee, Fiodar Kazhamiaka, Kiwan Maeng, Manoj Chakkaravarthy, Udit Gupta, David Brooks, and Carole-Jean Wu. Carbon explorer: A holistic approach for designing carbon aware datacenters. *Proceedings of the 28th ACM International Conference on Architectural Support for Programming Languages and Operating Systems*, 2023a.

Bilge Acun, Benjamin Lee, Fiodar Kazhamiaka, Kiwan Maeng, Udit Gupta, Manoj Chakkaravarthy, David Brooks, and Carole-Jean Wu. Carbon explorer: A holistic framework for designing carbon aware datacenters. In *Proceedings of the 28th ACM International Conference on Architectural Support for Programming Languages and Operating Systems, Volume 2*. ACM, January 2023b. doi: 10.1145/3575693.3575754. URL https://doi.org/10.1145/3575693.3575754.

Alibaba Group. Alibaba production cluster data. https://github.com/alibaba/clusterdata, 2017. Accessed: 2024-06-05.

Jyotika Athavale, Cullen Bash, Wesley Brewer, Matthias Maiterth, Dejan Milojicic, Harry Petty, and Soumyendu Sarkar. Digital twins for data centers. *Computer*, 57(10):151–158, 2024. doi: 10.1109/MC.2024.3436945.

Thomas J Breen, Ed J Walsh, Jeff Punch, Amip J Shah, and Cullen E Bash. From chip to cooling tower data center modeling: Part i influence of server inlet temperature and temperature rise across cabinet. In *2010 12th IEEE Intersociety Conference on Thermal and Thermomechanical Phenomena in Electronic Systems*, pages 1–10. IEEE, 2010.

Drury B Crawley, Linda K Lawrie, Curtis O Pedersen, and Frederick C Winkelmann. Energy plus: energy simulation program. *ASHRAE journal*, 42(4):49–56, 2000.

Data Center Map. Data center map: Directory of data centers. https://www.datacentermap.com/usa/. Accessed: 2024-06-10.

Christian Schroeder de Witt, Tarun Gupta, Denys Makoviichuk, Viktor Makoviychuk, Philip H. S. Torr, Mingfei Sun, and Shimon Whiteson. Is independent learning all you need in the starcraft multi-agent challenge?, 2020. URL https://arxiv.org/abs/2011.09533.

Google. Google cluster workload traces. https://github.com/google/cluster-data, 2019. Accessed: 2024-06-05.

Javier Jiménez-Raboso, Alejandro Campoy-Nieves, Antonio Manjavacas-Lucas, Juan Gómez-Romero, and Miguel Molina-Solana. Sinergym: A building simulation and control framework for training reinforcement learning agents. In *Proceedings of the 8th ACM International Conference on Systems for Energy-Efficient Buildings, Cities, and Transportation*, page 319–323, New York, NY, USA, 2021. Association for Computing Machinery. ISBN 9781450391146. doi: 10.1145/3486611.3488729. URL https://doi.org/10.1145/3486611.3488729.

Takao Moriyama, Giovanni De Magistris, Michiaki Tatsubori, Tu-Hoa Pham, Asim Munawar, and Ryuki Tachibana. Reinforcement learning testbed for power-consumption optimization. *CoRR*, abs/1808.10427, 2018. URL http://arxiv.org/abs/1808.10427.

Avisek Naug, Antonio Guillen, Ricardo Luna Gutiérrez, Vineet Gundecha, Sahand Ghorbanpour, Lekhapriya Dheeraj Kashyap, Dejan Markovikj, Lorenz Krause, Sajad Mousavi, Ashwin Ramesh Babu, and Soumyendu Sarkar. Pydcm: Custom data center models with reinforcement learning for sustainability. In *Proceedings of the 10th ACM International Conference on Systems for Energy-Efficient Buildings, Cities, and Transportation*, BuildSys '23, page 232–235, New York, NY, USA, 2023a. Association for Computing Machinery. ISBN 9798400702303. doi: 10.1145/3600100.3623732. URL https://doi.org/10.1145/3600100.3623732.

Avisek Naug, Antonio Guillen, Ricardo Luna Gutierrez, Vineet Gundecha, Sahand Ghorbanpour, Sajad Mousavi, Ashwin Ramesh Babu, and Soumyendu Sarkar. A configurable pythonic data center model for sustainable cooling and ml integration. In *NeurIPS 2023 Workshop on Tackling Climate Change with Machine Learning*, 2023b.

Soumyendu Sarkar, Vineet Gundecha, Alexander Shmakov, Sahand Ghorbanpour, Ashwin Ramesh Babu, Paolo Faraboschi, Mathieu Cocho, Alexandre Pichard, and Jonathan Fievez. Multi-agent reinforcement learning controller to maximize energy efficiency for multi-generator industrial wave energy converter. *Proceedings of the AAAI Conference on Artificial Intelligence*, 36(11): 12135–12144, Jun. 2022. doi: 10.1609/aaai.v36i11.21473. URL https://ojs.aaai.org/index.php/AAAI/article/view/21473.

Soumyendu Sarkar, Ashwin Ramesh Babu, Sajad Mousavi, Sahand Ghorbanpour, Vineet Gundecha, Ricardo Luna Gutierrez, Antonio Guillen, and Avisek Naug. Reinforcement learning based black-box adversarial attack for robustness improvement. In *2023 IEEE 19th International Conference on Automation Science and Engineering (CASE)*, pages 1–8. IEEE, 2023a.

Soumyendu Sarkar, Ashwin Ramesh Babu, Sajad Mousavi, Vineet Gundecha, Sahand Ghorbanpour, Alexander Shmakov, Ricardo Luna Gutierrez, Antonio Guillen, and Avisek Naug. Robustness with black-box adversarial attack using reinforcement learning. In *AAAI 2023: Proceedings of the Workshop on Artificial Intelligence Safety 2023 (SafeAI 2023)*, volume 3381. https://ceur-ws.org/Vol-3381/8.pdf, 2023b.

Soumyendu Sarkar, Antonio Guillen, Zachariah Carmichael, Vineet Gundecha, Avisek Naug, Ashwin Ramesh Babu, and Ricardo Luna Gutierrez. Enhancing data center sustainability with a 3d cnn-based cfd surrogate model. In *NeurIPS 2023 Workshop on Tackling Climate Change with Machine Learning*, 2023c.

Soumyendu Sarkar, Vineet Gundecha, Sahand Ghorbanpour, Alexander Shmakov, Ashwin Ramesh Babu, Avisek Naug, Alexandre Pichard, and Mathieu Cocho. Function approximation for reinforcement learning controller for energy from spread waves. In *IJCAI '23: Proceedings of the Thirty-Second International Joint Conference on Artificial Intelligence*, pages 6201–6209. Unknown publishers, August 2023d. ISBN 978-1-956792-03-4. doi: 10.24963/ijcai.2023/688.

Soumyendu Sarkar, Avisek Naug, Antonio Guillen, Ricardo Luna Gutierrez, Sahand Ghorbanpour, Sajad Mousavi, Ashwin Ramesh Babu, and Vineet Gundecha. Concurrent carbon footprint reduction (c2fr) reinforcement learning approach for sustainable data center digital twin. In *2023 IEEE 19th International Conference on Automation Science and Engineering (CASE)*, pages 1–8, 2023e. doi: 10.1109/CASE56687.2023.10260633.

Soumyendu Sarkar, Avisek Naug, Antonio Guillen, Ricardo Luna Gutierrez, Vineet Gundecha, Sahand Ghorbanpour, Sajad Mousavi, and Ashwin Ramesh Babu. Sustainable data center modeling: A multi-agent reinforcement learning benchmark. In *NeurIPS 2023 Workshop on Tackling Climate Change with Machine Learning*, 2023f.

Soumyendu Sarkar, Avisek Naug, Ricardo Luna Gutierrez, Antonio Guillen, Vineet Gundecha, Ashwin Ramesh Babu, and Cullen Bash. Real-time carbon footprint minimization in sustainable data centers with reinforcement learning. In *NeurIPS 2023 Workshop on Tackling Climate Change with Machine Learning*, 2023g.

Soumyendu Sarkar, Avisek Naug, Antonio Guillen, Ricardo Luna, Vineet Gundecha, Ashwin Ramesh Babu, and Sajad Mousavi. Sustainability of data center digital twins with reinforcement learning. *Proceedings of the AAAI Conference on Artificial Intelligence*, 38(21):23832–23834, Mar. 2024a. doi: 10.1609/aaai.v38i21.30580. URL https://ojs.aaai.org/index.php/AAAI/article/view/30580.

Soumyendu Sarkar, Avisek Naug, Ricardo Luna, Antonio Guillen, Vineet Gundecha, Sahand Ghorbanpour, Sajad Mousavi, Dejan Markovikj, and Ashwin Ramesh Babu. Carbon footprint reduction for sustainable data centers in real-time. *Proceedings of the AAAI Conference on Artificial Intelligence*, 38(20):22322–22330, Mar. 2024b. doi: 10.1609/aaai.v38i20.30238. URL https://ojs.aaai.org/index.php/AAAI/article/view/30238.

Paul Scharnhorst, Baptiste Schubnel, Carlos Fernández Bandera, Jaume Salom, Paolo Taddeo, Max Boegli, Tomasz Gorecki, Yves Stauffer, Antonis Peppas, and Chrysa Politi. Energym: A building model library for controller benchmarking. *Applied Sciences*, 11(8), 2021. ISSN 2076-3417. doi: 10.3390/app11083518. URL https://www.mdpi.com/2076-3417/11/8/3518.

John Schulman, Filip Wolski, Prafulla Dhariwal, Alec Radford, and Oleg Klimov. Proximal policy optimization algorithms. *arXiv preprint arXiv:1707.06347*, 2017.

Ratnesh Sharma, Amip Shah, Cullen Bash, Tom Christian, and Chandrakant Patel. Water efficiency management in datacenters: Metrics and methodology. In *2009 IEEE International Symposium on Sustainable Systems and Technology*, pages 1–6, 2009. doi: 10.1109/ISSST.2009.5156773.

Mohammed Shublaq and Ahmad K. Sleiti. Experimental analysis of water evaporation losses in cooling towers using filters. *Energy and Buildings*, 231:110603, 2020.

SPX Cooling Technologies. Water usage calculator, 2023. URL https://spxcooling.com/water-calculator/. Accessed: 2024-06-11.

Kaiyu Sun, Na Luo, Xuan Luo, and Tianzhen Hong. Prototype energy models for data centers. *Energy and Buildings*, 231:110603, 2021.

José R. Vázquez-Canteli, Jérôme Kämpf, Gregor Henze, and Zoltan Nagy. Citylearn v1.0: An openai gym environment for demand response with deep reinforcement learning. In *Proceedings of the 6th ACM International Conference on Systems for Energy-Efficient Buildings, Cities, and Transportation*, BuildSys '19, page 356–357, New York, NY, USA, 2019. Association for Computing Machinery. ISBN 9781450370059. doi: 10.1145/3360322.3360998. URL https://doi.org/10.1145/3360322.3360998.

Michael Wetter, Wangda Zuo, Thierry S Nouidui, and Xiufeng Pang. Modelica buildings library. *Journal of Building Performance Simulation*, 7(4):253–270, 2014.

Christopher Yeh, Victor Li, Rajeev Datta, Yisong Yue, and Adam Wierman. Sustaingym: A benchmark suite of reinforcement learning for sustainability applications. In *Thirty-seventh Conference on Neural Information Processing Systems Datasets and Benchmarks Track. PMLR*, page 1, 2023.

Chao Yu, Akash Velu, Eugene Vinitsky, Jiaxuan Gao, Yu Wang, Alexandre Bayen, and Yi Wu. The surprising effectiveness of PPO in cooperative multi-agent games. In *Thirty-sixth Conference on Neural Information Processing Systems Datasets and Benchmarks Track*, 2022. URL https://openreview.net/forum?id=YVXaxB6L2Pl.

Kun Zhang, David Blum, Hwakong Cheng, Gwelen Paliaga, Michael Wetter, and Jessica Granderson. Estimating ashrae guideline 36 energy savings for multi-zone variable air volume systems using spawn of energyplus. *Journal of Building Performance Simulation*, 15(2):215–236, 2022.

Yifan Zhong, Jakub Grudzien Kuba, Xidong Feng, Siyi Hu, Jiaming Ji, and Yaodong Yang. Heterogeneous-agent reinforcement learning. *Journal of Machine Learning Research*, 25(32): 1–67, 2024. URL http://jmlr.org/papers/v25/23-0488.html.

Wangda Zuo, Michael Wetter, James VanGilder, Xu Han, Yangyang Fu, Cary Faulkner, Jianjun Hu, Wei Tian, and Michael Condor. Improving Data Center Energy Efficiency Through End-to-End Cooling Modeling and Optimization. Final Report, April 2021. [Online; accessed 14. Oct. 2024].